\documentclass{article}

\PassOptionsToPackage{numbers,compress}{natbib}
\PassOptionsToPackage{table}{xcolor}

\newif\ifcamera
\cameratrue

\usepackage[final,eandd]{neurips_2026}

\usepackage[utf8]{inputenc}
\usepackage[T1]{fontenc}
\usepackage{url}
\usepackage{booktabs}
\usepackage{amsfonts}
\usepackage{amssymb}
\usepackage{amsmath}
\usepackage{nicefrac}
\usepackage{microtype}
\usepackage{xcolor}
\usepackage{graphicx}
\usepackage{multirow}
\usepackage{array}
\usepackage{tabularx}
\usepackage{makecell}
\usepackage{enumitem}
\usepackage{adjustbox}
\usepackage{hyperref}
\usepackage{titletoc}
\usepackage{wrapfig}
\usepackage{placeins}
\usepackage[most]{tcolorbox}
\usepackage{listings}
\usepackage{needspace}

\newcommand{\framework}{\textsc{MedOpenClaw}}
\newcommand{\bench}{\textsc{MedFlow-Bench}}
\newcommand{\copilot}{\textsc{MedCopilot}}

\definecolor{OpenUpBg}{HTML}{EAF7EA}
\definecolor{OpenDownBg}{HTML}{FCE8E6}
\definecolor{OpenFlatBg}{HTML}{F1F3F4}
\definecolor{OpenBlockBg}{HTML}{EEF3F8}
\definecolor{OpenSubBg}{HTML}{F7F9FC}
\definecolor{OpenSummaryBg}{HTML}{F4F4F4}

\definecolor{ToolUpBg}{HTML}{EAF7EA}
\definecolor{ToolDownBg}{HTML}{FCE8E6}
\definecolor{ToolFlatBg}{HTML}{F1F3F4}
\definecolor{ToolNA}{HTML}{FAFAFA}
\definecolor{ToolHeadBg}{HTML}{EEF3F8}

\definecolor{MainBlockBg}{HTML}{EEF3F8}
\definecolor{MainSummaryBg}{HTML}{F4F4F4}
\definecolor{MainStrictBg}{HTML}{FFF7E6}
\definecolor{MainLocBg}{HTML}{EAF3FF}
\definecolor{MainMissingBg}{HTML}{F1F3F4}

\definecolor{AppBlockBg}{HTML}{EEF3F8}
\definecolor{AppSummaryBg}{HTML}{F4F4F4}
\definecolor{AppDetailBg}{HTML}{F7F9FC}
\definecolor{AppStrictBg}{HTML}{FFF7E6}
\definecolor{AppLocBg}{HTML}{EAF3FF}
\definecolor{AppMissingBg}{HTML}{F1F3F4}

\newcolumntype{D}{>{\columncolor{AppDetailBg}}c}
\newcolumntype{S}{>{\columncolor{AppStrictBg}}c}
\newcolumntype{G}{>{\columncolor{AppLocBg}}c}

\definecolor{FailAccent}{HTML}{274C77}
\definecolor{FailFrame}{HTML}{CBD5E1}
\definecolor{FailHeaderBg}{HTML}{F8FAFC}
\definecolor{FailCodeBg}{HTML}{FAFBFC}
\definecolor{FailWarnBg}{HTML}{FFF8E5}
\definecolor{FailWarnFrame}{HTML}{E8D7A8}

\newtcolorbox{casecard}[1]{
  enhanced,
  breakable,
  colback=white,
  colframe=FailFrame,
  boxrule=0.45pt,
  arc=0.8pt,
  left=6pt,
  right=6pt,
  top=5pt,
  bottom=5pt,
  before skip=0.45em,
  after skip=0.45em,
  borderline west={2.0pt}{0pt}{FailAccent},
  title={#1},
  colbacktitle=FailHeaderBg,
  coltitle=black,
  fonttitle=\bfseries\small,
  boxed title style={
    colback=FailHeaderBg,
    colframe=FailHeaderBg,
    boxrule=0pt,
    sharp corners,
  },
  attach boxed title to top left={xshift=4pt,yshift=-1.8mm},
  top=7pt
}

\newtcblisting{tracecompact}[1]{
  enhanced,
  listing only,
  breakable,
  colback=FailCodeBg,
  colframe=FailFrame,
  boxrule=0.35pt,
  arc=0.6pt,
  left=4pt,
  right=4pt,
  top=3pt,
  bottom=3pt,
  before skip=0.35em,
  after skip=0.45em,
  title={#1},
  colbacktitle=white,
  coltitle=black,
  fonttitle=\bfseries\scriptsize,
  listing options={
    basicstyle=\ttfamily\tiny,
    breaklines=true,
    columns=fullflexible,
    keepspaces=true,
    showstringspaces=false
  }
}

\newtcolorbox{mechanismbox}{
  enhanced,
  breakable,
  colback=FailWarnBg,
  colframe=FailWarnFrame,
  boxrule=0.4pt,
  arc=0.8pt,
  left=6pt,
  right=6pt,
  top=5pt,
  bottom=5pt,
  before skip=0.45em,
  after skip=0.85em,
  borderline west={2.0pt}{0pt}{FailWarnFrame}
}

\newcommand{\toolup}[1]{\,{\scriptsize\ensuremath{(\uparrow #1)}}}
\newcommand{\tooldown}[1]{\,{\scriptsize\ensuremath{(\downarrow #1)}}}
\newcommand{\toolflat}{\,{\scriptsize\ensuremath{(\leftrightarrow 0.00)}}}

\newcommand{\tcup}[2]{\cellcolor{ToolUpBg}#1{\scriptsize\,\toolup{#2}}}
\newcommand{\tcdown}[2]{\cellcolor{ToolDownBg}#1{\scriptsize\,\tooldown{#2}}}
\newcommand{\tcflat}[1]{\cellcolor{ToolFlatBg}#1{\scriptsize\,\toolflat}}
\newcommand{\tcna}{\cellcolor{ToolNA}{\textemdash}}

\newcommand{\oalt}[1]{\cellcolor{OpenFlatBg}#1}

\title{\framework{} and \bench{}: \\ Auditing Medical Agents in Full-Study Workflows}

\ifcamera
\author{%
\normalfont
\textbf{Weixiang Shen}$^{1,2,3}$ \quad
\textbf{Chengzhi Shen}$^{1,2}$ \quad
\textbf{Yanzhu Hu}$^{1,2}$ \quad
\textbf{Che Liu}$^{4}$ \\
\textbf{Junde Wu}$^{5}$ \quad
\textbf{Jiayuan Zhu}$^{5}$ \quad
\textbf{Xiao Han}$^{3}$ \quad
\textbf{Zongyue Li}$^{3,9}$ \\
\textbf{Jingpei Wu}$^{3,9}$ \quad
\textbf{Min Xu}$^{6}$ \quad
\textbf{Daguang Xu}$^{7}$ \quad
\textbf{Yueming Jin}$^{8}$ \\
\textbf{Benedikt Wiestler}$^{1,2,9,\dagger}$ \quad
\textbf{Daniel Rueckert}$^{1,2,4,9,\dagger}$ \quad
\textbf{Jiazhen Pan}$^{1,2,9,\dagger}$ 
\\[0.65em]
$^1$Technical University of Munich (TUM) \quad
$^2$TUM University Hospital \quad
$^3$LMU Munich \\
$^4$Imperial College London \quad
$^5$University of Oxford \quad
$^6$Carnegie Mellon University \\
$^7$NVIDIA \quad
$^8$National University of Singapore \quad
$^9$Munich Center for Machine Learning \\
$^\dagger$Corresponding authors
}

\else
\author{Anonymous Authors \\
Paper under double-blind review}
\fi

\begin{document}

\maketitle

\vspace{-22pt}
\begin{abstract}
\vspace{-12pt}

Medical imaging benchmarks often evaluate VLMs on pre-selected 2D images, slices, crops, or patches, making evaluation closer to visual recognition. Real clinical workflows impose a different burden: readers must search through complete studies, operate imaging software, navigate across slices and magnifications, and document visual evidence that can be audited. We argue that this evidence-producing workflow is a critical missing evaluation axis for medical imaging agents. To study it, we introduce MedFlowBench, a full-study benchmark for VLM agents, together with MedOpenClaw, a controlled and replayable runtime in which agents operate medical imaging viewers such as 3D Slicer and QuPath. In each episode, an agent inspects a complete radiology study or whole-slide pathology image, returns a task answer, and submits structured evidence, including key slices, coordinates, regions of interest, or lesion-state fields. This evidence is automatically checked against withheld masks, annotations, and labels. Across evaluated models, final answer-only scoring gives an overly optimistic picture: when answers must also be supported by correct evidence, performance drops substantially on complex workflows. We further find that adding image-analysis tools does not by itself solve the problem. Tools help when they make a complex procedure simple and reliable, but agents still struggle when they must choose inputs, manage viewer state, and verify intermediate outputs over multiple steps. MedFlowBench exposes whether medical imaging agents can produce auditable evidence from complete studies, rather than plausible answers from selected images.
\vspace{-8pt}

\end{abstract}

\vspace{-8pt}
\section{Introduction}
\vspace{-10pt}
Medical vision-language models have progressed rapidly, from early medical VQA benchmarks \cite{lau2018vqarad,liu2021slake} to more recent medical multimodal systems and expert-level evaluation sets \cite{chen2024huatuo,zuo2025medxpertqa,medxpertqa_hf}. Yet much of their medical imaging evaluation still relies on a simplified proxy setting: the model is given one or a few \textbf{pre-selected diagnostically relevant 2D images} and asked to answer a localized question. This setup is useful for testing recognition on selected inputs, but it removes the central difficulty of medical imaging, especially radiology, where findings must often be found across a complete volumetric study. A parallel abstraction appears in pathology when whole-slide images are reduced to selected patches or slide-level labels. It also keeps the decision process largely opaque: the model returns an answer, but not a replayable account of where it looked, what evidence it gathered, or how the final conclusion was reached.

This creates a substantial gap to real clinical workflow \citep{coiera2019lastmile,vandesande2024multifaceted,korfiatis2025workflow}. In practice, medical imaging analysis is a study-level and software-mediated process. A reader must inspect a full 3D examination, choose relevant series or modalities, navigate across many slices, adjust display settings such as windowing or fusion, compare evidence across views and timepoints, and often perform measurements, registration, segmentation, or specialized analysis before committing to an interpretation. In whole-slide pathology, the corresponding workflow requires opening a gigapixel slide, panning and zooming across tissue regions, and recording evidence at appropriate coordinates and magnifications. Many clinically relevant findings are not visible in a single image. They emerge only across adjacent slices, across sequences, across longitudinal studies, or after the relevant region has been localized \citep{desmet2006twoslice,qureshi2016dcect,li2023cine}. A meaningful evaluation setting should therefore test not only whether an agent can produce the correct answer, but whether it can search complete studies, maintain the relevant viewer state, and leave evidence that is transparent, replayable, and auditable.

To evaluate this setting, we introduce \textbf{\framework{}} and \textbf{\bench{}}. \framework{} is an auditable runtime that lets a backbone VLM operate standard medical imaging viewers, including 3D Slicer~\citep{fedorov2012slicer} for volumetric radiology and QuPath~\citep{bankhead2017qupath} for whole-slide pathology. It restricts the agent to predefined software actions for viewer navigation, evidence capture, and selected analysis operations, and records each step as a replayable trace. Built on this runtime, \bench{} evaluates full-study medical imaging episodes rather than pre-selected inputs. It covers five task families across radiology and pathology: brain tumor subtype assessment on pre-operative multi-sequence MRI, longitudinal treatment-response assessment on MRI, lung PET/CT analysis, breast histology classification, and lymph-node metastasis assessment on whole-slide images. These tasks span clinically meaningful workflows that require lesion localization, integration of multimodal or lesion-status labels for longitudinal MRI tasks, gigapixel slide navigation, and structured evidence-grounded reasoning.

In \bench{}, each agent must inspect a complete study and return both a task answer and task-specific structured evidence. This evidence is not a free-text rationale; it consists of fields that tie the answer back to image content, such as key-slice positions, lesion or slide coordinates, regions of interest, or lesion-status labels for longitudinal MRI tasks. The benchmark checks these fields with deterministic rules using withheld ground-truth masks, annotations, or labels. This design evaluates whether an agent can not only produce the correct answer, but also identify and document the image evidence needed to support it.

Our results show that full-study evaluation is feasible for VLM agents: models can navigate medical viewers, inspect complete volumetric or whole-slide cases, and solve nontrivial study-level tasks without pre-selected inputs. However, final-answer-only scoring substantially overestimates workflow competence in realistic imaging workflows. When the answer must also be supported by correct evidence, scores drop markedly on complex tasks. These drops motivate evidence-grounded evaluation as a crucial benchmark requirement: medical imaging agents should be evaluated not only on task accuracy, but also on whether they provide correct, checkable evidence.

We further test whether specialized image-analysis operations inside the viewer close this evidence-grounding gap. They do not by themselves. Operations that collapse a complex procedure into a single well-scoped call can help; for example, MONAI-based VISTA3D~\citep{cardoso2022monai,he2025vista3d} provides automated volumetric segmentation and can improve localization in some PET/CT cases. In contrast, workflows that require the agent to compose several dependent software steps, such as interactive segmentation and registration, remain brittle. The resulting errors are software-workflow failures rather than only visual-recognition failures: agents may lose track of image space, bind outputs to the wrong viewer state, choose inappropriate parameters, or fail to verify the generated artifact. These failures motivate evidence-grounded full-study benchmarks.

Our contributions are:

\begin{itemize}[leftmargin=1.3em, topsep=1.0pt, itemsep=2.0pt, parsep=0pt, partopsep=0pt]

    \item We introduce \framework{}, an auditable runtime that enables VLM agents to operate professional medical software, including 3D Slicer and QuPath, on full studies rather than pre-selected diagnostically relevant inputs.
    \item We introduce \textbf{\bench{}}, a full-study benchmark for medical imaging reasoning across radiology and pathology. \bench{} evaluates not only final answers but also checkable evidence, including key slices, coordinates, regions, and longitudinal evidence.
   \item Using \bench{}, we show that current VLM agents can operate medical imaging software and solve some full-study radiology and pathology tasks without a pre-selection process. However, evaluating only the final answer overestimates their workflow competence. When evaluation also requires supporting evidence, performance drops markedly on complex tasks. We further show that access to specialized image-analysis tools is not sufficient for workflow competence. Automation tools can help when they reduce intermediate decision-making, but multi-step segmentation and registration workflows remain unreliable. 
\end{itemize}

\begin{figure*}[t]
    \centering
    \includegraphics[width=1\linewidth]{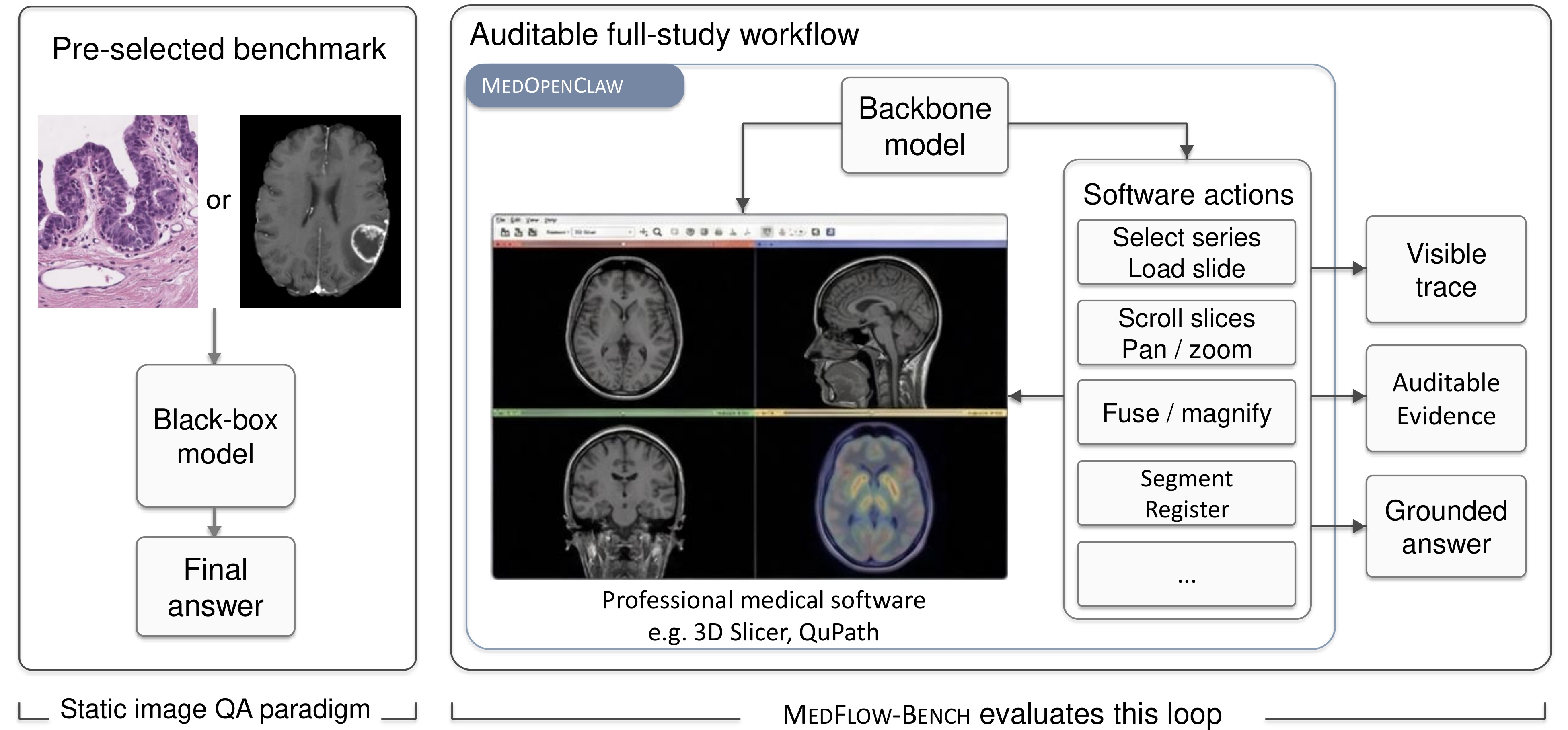}
   \caption{\textbf{Overview of \framework{} and \bench{}.}
\textbf{Left:} Conventional medical VQA benchmarks reduce evaluation to pre-selected 2D images and black-box final answers.
\textbf{Right:} \framework{} enables backbone models to operate professional medical software through bounded software actions, producing visible traces, auditable evidence, and grounded answers.
\bench{} evaluates this full-study interaction loop across radiology and whole-slide pathology workflows, including viewer-native navigation and specialized segmentation or registration functions.}
    \label{fig:main}
\end{figure*}

\section{Related Work}
\vspace{-12pt}
\noindent\textbf{Static medical VQA and medical VLM benchmarks.}
Early medical VQA datasets established language-conditioned evaluation on medical images
\cite{lau2018vqarad,he2020pathvqa,hu2024omnimedvqa,liu2021slake,zhang2023pmc}.
Subsequent medical QA and multimodal benchmarks broadened task scope, reasoning difficulty, and clinical coverage
\cite{chen2024huatuo,zuo2025medxpertqa,medxpertqa_hf,wang2025gpt5medical,kim2025benchmarking,yao2026medical,moll2025evaluating}.
Related lines such as medical report generation and image-grounded medical question answering also largely assume fixed image inputs rather than study-level interaction
\cite{fallahpour2025medrax,bercea2025nova,deng2025med,wang2025cxpmrg}.
Together, these benchmarks measure visual recognition, medical knowledge use, and language-conditioned reasoning on selected inputs, but they typically start from one or a few pre-selected diagnostic views rather than a full imaging study.
Patch- or slide-level pathology evaluations similarly reduce the search problem before the agent must navigate the whole slide.
We target this gap by evaluating auditable study-level reasoning over complete radiology and pathology studies.

\noindent\textbf{Medical agents.}
Another line of work studies medical agents that perform multi-step reasoning~\cite{hu2025landscape,kim2024mdagents,pan2025medvlm,wu2025medcasereasoning} and evidence gathering~\cite{wu2024medical,zhu2025ask,zhao2026agentic}. Some focus
on radiology-oriented agents, especially for chest X-ray analysis or reporting
\cite{fallahpour2025medrax,sharma2024cxragent}, while others study broader
multimodal medical agents that select among specialized tools or APIs across tasks
\cite{li2024mmedagent,wang2025medagent}.
These systems make the reasoning loop more explicit than standard static benchmarks, but many still
operate on fixed images, isolated APIs, or abstracted tool interfaces rather than continuous
interaction with a viewer over a full tomographic exam.

\noindent\textbf{Full-study and interactive medical imaging systems.}
More closely related work begins to address study-level or interactive radiology settings. One line
studies volumetric or 3D medical image reasoning with language-guided analysis over image volumes
\cite{baharoon2025rexgroundingct,hoopes2024voxelprompt,mao2025ctagent}. Another line evaluates agents in
simulated or simplified radiology environments
\cite{zheng2024radabench,jiang2025medagentbench}. A third line explores
natural-language assistance or copilot-style interaction inside existing imaging software such as
3D Slicer \cite{barr2024slicerchat,namic2025copilot}. These directions move closer to
clinical imaging workflow, but differ in whether the main emphasis is on volumetric reasoning,
environment design, or software-integrated interaction. Our work sits at the intersection of these
threads by focusing on study-level evaluation in a real viewer with preserved execution traces.

\noindent\textbf{General-purpose agent runtimes.}
At the systems level, \framework{} is related to real-environment agent benchmarks for software engineering, desktop use, and tabular tasks~\cite{jimenez2024swebench,xie2024osworld,erickson2025tabarena}, as well as general-purpose runtimes such as OpenClaw~\cite{openclaw2026,openclaw_tools2026,openclaw_policy2026,openclaw_sandbox2026}.
Unlike open-ended agent runtimes, \framework{} builds bounded control into the medical-imaging interface itself: agents are restricted to software-native actions and vetted advanced operations under a standardized protocol.
This makes traces safer, replayable, and aligned with clinical evidence requirements.

\section{\framework{} Runtime}

\framework{} defines the software-native interaction contract used by \bench{}. It sits between a backbone VLM agent and standard medical imaging environments, including 3D Slicer~\citep{fedorov2012slicer} for volumetric radiology studies and QuPath~\citep{bankhead2017qupath} for whole-slide pathology images. We use these environments because full-study evidence is naturally represented through software state: selected series, slice index, viewport, windowing or fusion settings, Right-Anterior-Superior (RAS) coordinates, WSI coordinates, masks, measurements, segmentations, registrations, and saved artifacts. These forms of evidence are difficult to represent or evaluate from static selected images alone. \framework{} runs externally without modifying the viewer source code and exposes a fixed interface for study inspection, evidence capture, and analysis. Through this interface, the agent can select series, scroll through slices, adjust display settings, bookmark views, take measurements, pan and zoom whole-slide images, convert viewer points into image coordinates, and export evidence. Figure~\ref{fig:main} summarizes this contract.

The exposed action space is grouped into three roles: Viewer Control, Evidence Capture, and Analysis Operators. \textbf{Viewer Control} actions manipulate the visible study state, such as series selection, slice scrolling, panning, zooming, windowing, and basic foreground/background fusion of already loaded or prealigned volumes. \textbf{Evidence Capture} actions save checkable evidence from the current state, including bookmarked views, key-slice or coordinate locations, drawn regions or masks, and measurement logs. \textbf{Analysis Operators} invoke functions that create derived study state, such as segmentation, registration, resampling, quantitative analysis, or automated expert-model outputs. MONAI-based operation packs~\citep{cardoso2022monai,diazpinto2024monailabel} and 3D VISTA~\citep{he2025vista3d} are exposed as Analysis Operators through 3D Slicer. Section~\ref{sec:bench} describes how \bench{} maps these actions to task protocols and scoring rules.

Implementation details of the REST interface, named bridge handlers, QuPath operation registry, and safety restrictions are provided in Appendix~\ref{app:runtime_interface}. The bounded design makes auditability operational rather than implicit. The runtime logs each action invocation together with its arguments, the resulting viewer-state snapshot, and any generated artifacts. These records allow the execution to be reconstructed after the fact, including which views were accessed, which operations were executed, and what evidence was available when the final answer was produced.

Although \bench{} uses \framework{} for autonomous agent evaluation, the same traceable interaction contract can also support human-in-the-loop interfaces. We implement \copilot{} as an example application built on top of \framework{}, where generated actions and evidence artifacts remain visible for clinician inspection. We treat this as a demonstration of traceable interaction rather than evidence of clinical deployment or workflow-efficiency gains.

\section{\bench{}: Study-Level Evaluation}\label{sec:bench}
Building on the runtime contract above, \bench{} defines the evaluation distribution: the clinical task families, allowed tracks, canonical answer schemas, and evidence-based scoring rules. \bench{} evaluates study-level and whole-slide reasoning rather than perception from pre-selected 2D images. It is designed around full-study interactive access, cross-modality and longitudinal reasoning, active evidence acquisition, differential diagnosis, and agentic execution in professional medical software. A benchmark episode is defined at the study or slide level rather than the image-patch level. Each episode specifies (i) a case package containing the full volumetric exam or whole-slide image and metadata, (ii) the software environment used by the agent, (iii) a task prompt that asks for a case-level or slide-level decision, (iv) an allowed action space determined by the evaluation track, and (v) a canonical answer and evidence schema used for scoring.

\begin{figure*}[t]
    \centering
    \includegraphics[
        width=0.98\textwidth,
        height=0.70\textheight,
        keepaspectratio
    ]{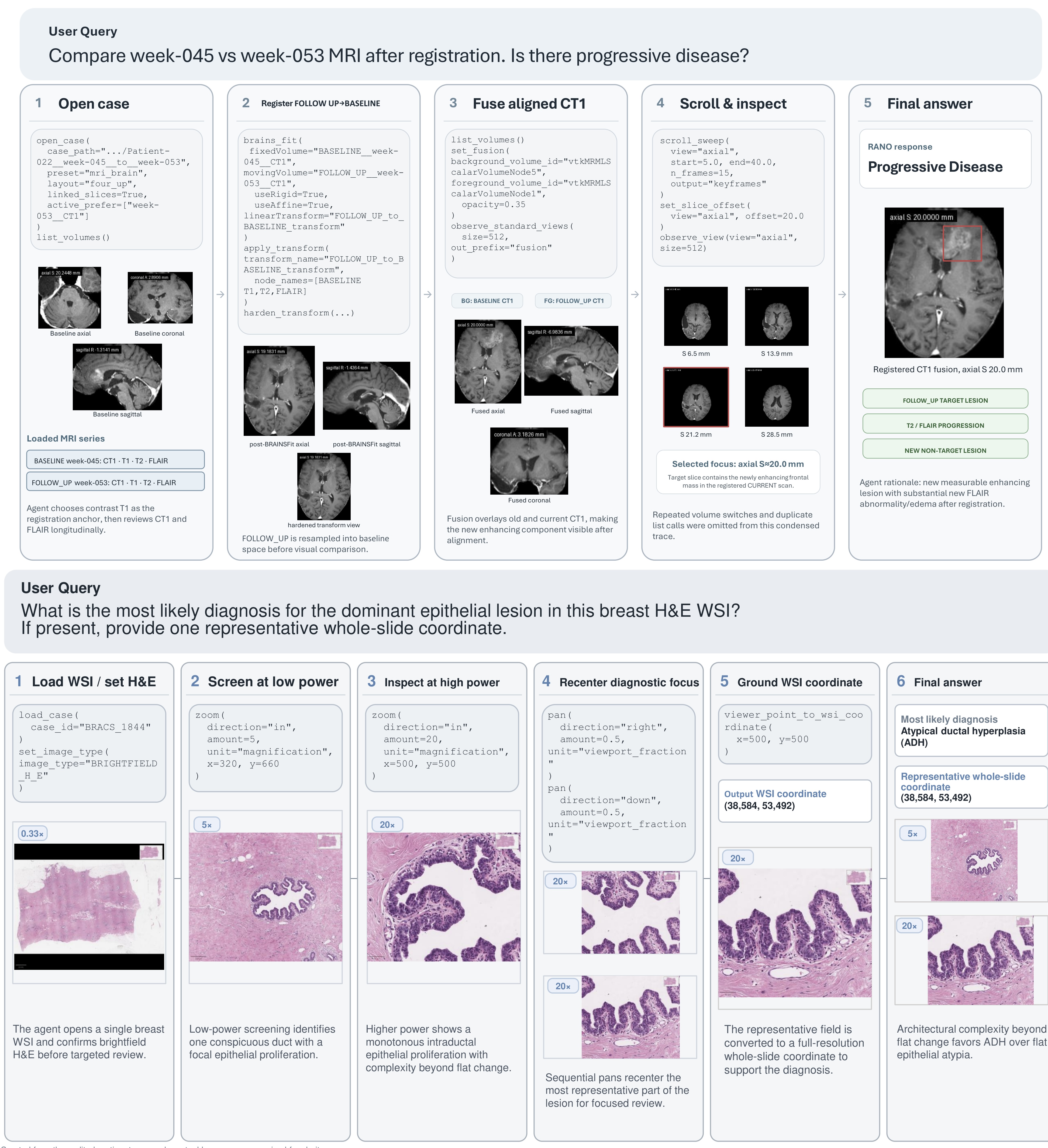}
    \vspace{-3pt}
    \caption{\textbf{Representative auditable execution traces from \framework{} in brain MRI and breast WSI workflows.}
    \textbf{Top:} For longitudinal brain MRI, the agent loads baseline and follow-up studies, registers the week-053 scan to week-045, inspects fused aligned volumes, and reports RANO~\citep{wen2010rano} progressive disease.
    \textbf{Bottom:} For breast H\&E WSI, the agent screens the slide, reviews the dominant epithelial lesion at higher magnification, maps the diagnostic field to a whole-slide coordinate, and reports atypical ductal hyperplasia.
    The traces are compressed from longer runtime logs; full records retain operation calls, arguments, viewer states, image evidence, locations, and final outputs for auditability.}
    \label{fig:tooltrace}
    \vspace{-8pt}
\end{figure*}

\vspace{-10pt}
\noindent\paragraph{Example benchmark episode.} As a concrete example, Figure~\ref{fig:tooltrace} illustrates abbreviated, decision-relevant execution traces from the Longitudinal MRI and Breast WSI modules. In the longitudinal MRI scenario, the agent loads the BASELINE and FOLLOW-UP studies, registers the follow-up CT1 volume to the baseline CT1 volume, inspects fused aligned views across timepoints, and reports a RANO response label with FOLLOW-UP-study evidence fields. In the breast WSI scenario, the agent screens the slide, zooms into a diagnostic epithelial region, converts the viewed field to whole-slide coordinates, and returns the slide-level diagnosis with coordinate-grounded evidence. Across both examples, the query, operation calls, visual outputs, and response remain externally inspectable instead of collapsing into an opaque intermediate state.

\noindent\paragraph{Task families and evidence.}

\vspace{-10pt}
\bench{} covers five challenging diagnostic imaging tasks across radiology and pathology: brain tumor subtype assessment on pre-operative MRI, longitudinal treatment-response assessment on brain MRI, lung PET/CT analysis, breast histology classification, and lymph-node metastasis WSI assessment. These tasks were chosen to span representative, clinically meaningful workflows in which diagnosis requires localization, integration of multimodal or longitudinal evidence, and structured clinical reasoning. In radiology, pre-operative brain tumor MRI emphasizes intracranial lesion localization, cross-sequence integration, anatomical reasoning, and tumor subtype assessment~\citep{calabrese2022ucsfpdgm}, while longitudinal response assessment requires RANO-style comparison of baseline and follow-up studies, including tracking target and non-target lesions over time~\citep{suter2022lumiere}. Lung PET/CT broadens the radiology setting to thoracic oncology, requiring joint interpretation of metabolic (PET) and anatomical (CT) findings for tumor localization, staging-relevant assessment, histology, and grade~\citep{bakr2018nsclc}. In pathology, the WSI tasks evaluate gigapixel slide analysis with localized diagnostic evidence: breast WSI covers benign, atypical, and malignant breast lesion patterns~\citep{brancati2022bracs}, while lymph-node WSI focuses on metastasis detection and nodal-status assessment within TNM staging, a clinically important component of cancer staging~\citep{bandi2019camelyon17}. Across all tasks, agents must operate professional medical imaging software and provide structured, verifiable evidence whenever required by the task definition.

\vspace{-10pt}
\noindent\paragraph{The Three-Track Design.}
To support both model evaluation and systems work, \bench{} separates the solution space into three tracks rather than mixing all methods on one leaderboard. All tracks use identical cases, task formulations, and metrics. Conceptually, Track C is a diagnostic compatibility track, whereas Track B is the main software-use intervention.

\begin{itemize}[leftmargin=1.3em, topsep=2pt, itemsep=2pt]
    \item \textbf{Track A: Viewer-Native.} A test of viewer-native study inspection without specialized analysis functions. Aligning with the first layer of our runtime architecture, methods use \framework{} to drive the relevant medical viewer using only primitive viewer-native actions, such as series selection, scrolling, windowing, and basic prealigned foreground/background fusion in 3D Slicer, or slide loading, panning, zooming, and viewport capture in QuPath. By excluding specialized analysis functions, this track focuses on visual search, slice-to-slice or region-to-region synthesis, and evidence acquisition rather than configuring specialized functions.
    \item  \textbf{Track B: Advanced Operations.} The main systems track, allowing access to advanced operations via \framework{} where such operations are defined. Here, \textit{advanced operations} refer to specialized functions beyond viewer-native inspection that require the agent to invoke, configure, and verify algorithm- or model-backed functionality inside professional medical imaging software, including segmentation operators, registration workflows, and automated expert-model workflows such as 3D VISTA. This track tests whether a model can decide when such an operation is needed, set parameters, preserve the returned artifacts in software state, and integrate them back into the diagnostic trajectory.
    \item \textbf{Track C: Runtime-agnostic.} Methods are not required to use \framework{} and may use any alternative pipeline that consumes the raw cases and outputs the canonical answer and evidence schema. We include this track to ensure the benchmark remains a universal standard rather than just a test of our specific runtime. This leaves room for future full-study paradigms.
\end{itemize}
\vspace{-10pt}
\noindent\paragraph{Scoring metrics.}
We report three metric types. \textbf{Task} is the module primary answer score.
For LUMIERE, UCSF-PDGM, and BRACS, \textbf{Task} is single-label accuracy.
For CAMELYON17, \textbf{Task} is exact-match accuracy over tumor presence and
largest-metastasis category. For NSCLC, \textbf{Task} is the partial score
\#correct/5 over tumor location, pathological T stage, pathological N stage,
histology, and grade. \textbf{Strict} is the evidence-constrained version of
\textbf{Task}: if the module-specific evidence gate passes, \textbf{Strict}
equals \textbf{Task}; otherwise \textbf{Strict} is 0. Evidence gates are
deterministic and may use key-slice offsets, source-series provenance,
lesion-status fields, RAS points, or WSI coordinates depending on the module.
\textbf{Loc.} is reported separately when spatial ground truth is available
and measures coordinate localization under the module-specific denominator.

\section{Experiments}

All methods are evaluated under the task-specific MCQ or structured-answer protocol and the evidence requirements defined in Section~\ref{sec:bench}. Unless otherwise noted, the three tracks use identical cases, task formulations, and metrics.

\vspace{-10pt}
\noindent\paragraph{Viewer-Native Radiology Baselines.} Table~\ref{tab:main_results} reports the main 3D Slicer viewer-native results, with lesion-state and subtask-level details in Appendix Table~\ref{tab:main_results_detailed}. Overall, the table shows that current models still have limited ability to operate full-study radiology software with reliable evidence. Brain MRI is the most favorable setting: because the target tumor is relatively spatially salient, frontier proprietary models already achieve high localization and evidence-grounded performance, with Gemini 3.1 Pro Preview reaching 0.73 Task, 0.73 Strict, and 0.96 Loc. However, performance drops sharply in workflows that require longitudinal comparison or precise localization of smaller thoracic lesions. On Longitudinal MRI, even the best Strict score is only 0.18, indicating that correct response assessment with valid evidence remains difficult. On Lung PET/CT, the best Strict score is only 0.14, and the best localization score is 0.26, showing that models often fail to provide the precise RAS evidence needed for small-lesion thoracic oncology cases. Medical-domain models and smaller open-weight models are weaker still: their final-task scores are modest, and their Strict and localization scores are often near zero. Thus, viewer-native access enables some full-study reasoning, especially for salient Brain MRI findings, but reliable software-state navigation, evidence acquisition, and localization remain unresolved in clinically harder radiology workflows.

\begin{table*}[t]
    \centering
    {\scriptsize
    \setlength{\tabcolsep}{2.7pt}
    \renewcommand{\arraystretch}{1.12}
    \begin{adjustbox}{max width=\textwidth}
    \begin{tabular}{@{}>{\raggedright\arraybackslash}p{1.62in}
                    c >{\columncolor{MainStrictBg}}c
                    c >{\columncolor{MainStrictBg}}c >{\columncolor{MainLocBg}}c
                    c >{\columncolor{MainStrictBg}}c >{\columncolor{MainLocBg}}c
                    c >{\columncolor{MainStrictBg}}c >{\columncolor{MainLocBg}}c@{}}
        \toprule
        \multirow{2}{*}{Model}
            & \multicolumn{2}{c}{Longitudinal MRI}
            & \multicolumn{3}{c}{Brain MRI}
            & \multicolumn{3}{c}{Lung PET/CT}
            & \multicolumn{3}{c}{Model Avg.} \\
        \cmidrule(lr){2-3}\cmidrule(lr){4-6}\cmidrule(lr){7-9}\cmidrule(l){10-12}
            & Task & Strict
            & Task & Strict & Loc.
            & Task & Strict & Loc.
            & Task & Strict & Loc. \\
        \midrule

        \rowcolor{MainBlockBg}
        \multicolumn{12}{@{}l}{\textbf{General-purpose models}} \\
        GPT-5.4 mini~\citep{openai2026gpt54mini}
            & 0.23 & 0.05
            & 0.38 & 0.18 & 0.37
            & 0.31 & 0.00 & 0.00
            & 0.31 & 0.08 & 0.19 \\
        GPT-5.5~\citep{openai2026gpt55systemcard}
            & 0.33 & 0.04
            & 0.71 & 0.66 & 0.91
            & \textbf{0.55} & 0.03 & 0.07
            & \textbf{0.53} & 0.24 & 0.49 \\
        Gemini 3 Flash Preview~\citep{googledeepmind2025gemini3flashcard}
            & 0.30 & 0.07
            & 0.67 & 0.59 & 0.87
            & 0.45 & 0.13 & \textbf{0.26}
            & 0.47 & 0.26 & 0.57 \\
        Gemini 3.1 Pro Preview~\citep{googledeepmind2026gemini31procard}
            & \textbf{0.37} & 0.08
            & \textbf{0.73} & \textbf{0.73} & \textbf{0.96}
            & 0.50 & \textbf{0.14} & 0.21
            & \textbf{0.53} & \textbf{0.32} & \textbf{0.59} \\
        Claude Sonnet 4.6~\citep{anthropic2026claudesonnet46}
            & 0.33 & 0.17
            & 0.38 & 0.18 & 0.36
            & 0.26 & 0.00 & 0.00
            & 0.32 & 0.12 & 0.18 \\
        Claude Opus 4.7~\citep{anthropic2026claudeopus47}
            & 0.28 & \textbf{0.18}
            & 0.37 & 0.11 & 0.68
            & 0.43 & 0.01 & 0.01
            & 0.36 & 0.10 & 0.35 \\
        Qwen3.5-35B-A3B~\citep{qwen2026qwen35collection}
            & 0.33 & \textbf{0.18}
            & 0.53 & 0.15 & 0.11
            & 0.35 & 0.01 & 0.01
            & 0.40 & 0.11 & 0.06 \\
        Qwen3.5-27B~\citep{qwen2026qwen35collection}
            & 0.21 & 0.09
            & 0.53 & 0.27 & 0.36
            & 0.37 & 0.00 & 0.00
            & 0.37 & 0.12 & 0.18 \\
        Qwen3.5-9B~\citep{qwen2026qwen35collection}
            & 0.17 & 0.08
            & 0.35 & 0.05 & 0.06
            & 0.16 & 0.01 & 0.01
            & 0.23 & 0.05 & 0.04 \\
        Qwen3.5-4B~\citep{qwen2026qwen35collection}
            & 0.22 & 0.11
            & 0.32 & 0.06 & 0.03
            & 0.20 & 0.00 & 0.00
            & 0.25 & 0.06 & 0.02 \\
        Gemma 3 27B IT~\citep{gemmateam2025gemma3}
            & 0.32 & 0.01
            & 0.33 & 0.11 & 0.07
            & 0.35 & 0.00 & 0.00
            & 0.33 & 0.04 & 0.03 \\
        \rowcolor{MainSummaryBg}
        \emph{Mean over reported}
            & 0.28 & 0.10
            & 0.48 & 0.28 & 0.43
            & 0.36 & 0.03 & 0.05
            & 0.37 & 0.14 & 0.25 \\

        \addlinespace[0.15em]
        \rowcolor{MainBlockBg}
        \multicolumn{12}{@{}l}{\textbf{Medical-domain models}} \\
        MedGemma 1.5 4B IT~\citep{sellergren2025medgemma}
            & 0.16 & 0.02
            & 0.45 & 0.09 & 0.00
            & 0.16 & 0.00 & 0.00
            & 0.26 & 0.04 & 0.00 \\
        MedGemma 27B IT~\citep{sellergren2025medgemma}
            & 0.11 & 0.00
            & 0.30 & 0.00 & 0.03
            & 0.16 & 0.00 & 0.00
            & 0.19 & 0.00 & 0.02 \\
        Lingshu-32B~\citep{xu2025lingshu}
            & 0.16 & 0.00
            & 0.36 & 0.08 & 0.02
            & 0.31 & 0.00 & 0.00
            & 0.27 & 0.03 & 0.01 \\
        \rowcolor{MainSummaryBg}
        \emph{Mean over reported}
            & 0.14 & 0.01
            & 0.37 & 0.06 & 0.02
            & 0.21 & 0.00 & 0.00
            & 0.24 & 0.02 & 0.01 \\

        \bottomrule
    \end{tabular}
    \end{adjustbox}
    }
\caption{\textbf{Main viewer-native radiology results.}
    All rows use 3D Slicer viewer-native access.
    Yellow columns report \textbf{Strict}, the answer-plus-evidence score defined in the scoring protocol: credit is given only when the task answer is correct and the required evidence fields pass the task-specific deterministic checks; blue columns report localization where available.
    \textbf{Model Avg.} reports unweighted averages over task-family metrics: Task averages Longitudinal Task, Brain Task, and Lung PET/CT Task; Strict averages the three Strict columns; Loc. averages Brain MRI and Lung PET/CT localization.
    Detailed lesion-status and subtask-level metrics are reported in Appendix Table~\ref{tab:main_results_detailed}. Rows are grouped by model category, distinguishing general-purpose and medical-domain models.}
    \label{tab:main_results}
\end{table*}

\noindent\paragraph{Viewer-Native Pathology Baselines.}
\begin{wraptable}[12]{r}{0.50\textwidth}
    \vspace{-1.15em}
    \centering
    {\scriptsize
    \setlength{\tabcolsep}{1.8pt}
    \renewcommand{\arraystretch}{1.04}
    \begin{adjustbox}{max width=\linewidth}
    \begin{tabular}{@{}l@{\hspace{1.05em}}ccc@{\hspace{0.75em}}ccc@{}}
        \toprule
        \multirow{2}{*}{Model}
            & \multicolumn{3}{c}{\makecell{Breast\\WSI}}
            & \multicolumn{3}{c}{\makecell{Lymph-node\\WSI}} \\
        \cmidrule(lr){2-4}\cmidrule(l){5-7}
            & Task & Strict. & Loc.
            & Task & Strict. & Loc. \\
        \midrule
        GPT-5.4 mini
            & 0.20 & 0.04 & 0.03
            & 0.16 & 0.05 & \textbf{0.39} \\
        GPT-5.4
            & 0.16 & 0.10 & 0.07
            & \textbf{0.35} & 0.11 & 0.19 \\
        Gemini 3 Flash Preview
            & 0.20 & \textbf{0.16} & 0.13
            & 0.18 & 0.05 & 0.07 \\
        Gemini 3.1 Pro Preview
            & \textbf{0.34} & 0.15 & \textbf{0.19}
            & 0.26 & \textbf{0.12} & 0.38 \\
        \bottomrule
    \end{tabular}
    \end{adjustbox}
    }

    \vspace{-0.25em}
    \begin{minipage}{0.50\textwidth}
    \caption{\textbf{Viewer-native pathology.} Coordinate-grounded QuPath WSI scoring; Strict denotes answer-plus-evidence score and Loc.\ denotes coordinate localization where required.}
    \label{tab:viewer_pathology_filled}
    \end{minipage}
\end{wraptable}
Table~\ref{tab:viewer_pathology_filled} reports viewer-native WSI results under the same auditable-evidence requirement used throughout the benchmark. In pathology, the relevant software-native workflow is whole-slide search in QuPath, where the agent must identify a diagnostic region rather than only produce a slide-level label. The results follow the same high-level pattern as radiology: models can produce plausible case-level answers, but evidence and coordinate localization remain weak. This supports the generality of software-native evaluation while showing that verifiable diagnostic-region grounding in WSI remains unresolved. Scoring definitions are provided in Appendix~\ref{app:benchmark_modules_scoring}.

\vspace{-10pt}
\noindent\paragraph{Track C: Runtime-agnostic.}
Track C asks whether the viewer-native findings are merely an artifact of using \framework{} rather than a more general property of full-study evaluation. The full comparison is reported in Appendix~\ref{app:runtime_agnostic_track}, Table~\ref{tab:open_method}. The 2D slice-montage baseline compresses each volume into a static tiled overview, as illustrated in Appendix Figure~\ref{fig:slice_montage}, while native 3D VLMs such as M3D and RadFM consume volumetric inputs directly~\cite{bai2024m3d,wu2025radfm}. The results change some final-answer scores but do not yield a consistently stronger pipeline. In particular, 2D slice montages are poorly suited to certain tasks, such as longitudinal comparison, because they remove interactive cross-timepoint navigation and fusion; they also do not scale naturally to more advanced imaging analyses such as segmentation or registration. At the aggregate level, frontier models such as GPT-5.5 and Gemini 3.1 Pro Preview operating 3D Slicer substantially outperform native medical 3D VLM baselines such as M3D and RadFM on model-averaged Task, Strict, and localization metrics. Thus, Track C supports a narrow but important conclusion: \bench{} is not tied to one runtime interface, and the evidence-grounding bottleneck persists across runtime-agnostic alternatives. We report the full table in the appendix.
\vspace{-8pt}
\noindent\paragraph{Track B: Advanced Operations.} Having established that bypassing the runtime does not remove the evidence-grounding bottleneck, Track B tests the more consequential systems question: whether richer software operations help agents execute evidence-grounded imaging workflows. Table~\ref{tab:tooluse_trackb} separates clinical utility from agent executability. Segmentation and registration are valuable operations in real imaging workflows, but exposing them to current agents does not reliably improve benchmark performance: their gains are local and inconsistent, and the Task, Strict, and localization scores often remain flat or slightly decline. This points to a software-use bottleneck rather than a limitation of the operations themselves. These workflows require correct source-volume selection, spatial input, parameterization, state tracking, and output verification; current agents often fail at these steps, as detailed in Section~\ref{sec:software_use_failure}. By contrast, MONAI-based 3D VISTA~\citep{cardoso2022monai,he2025vista3d} is highly automated, deep-learning-based, and easier to invoke. On Lung PET/CT, agents use it more effectively, yielding consistent localization gains and model-dependent Strict improvements. Thus, advanced operations are useful when they reduce workflow burden, but complex professional operations remain constrained by the agent's software-control competence.

\begin{table*}[t]
    \centering
    {\scriptsize
    \setlength{\tabcolsep}{2.7pt}
    \renewcommand{\arraystretch}{1.13}
    \begin{adjustbox}{max width=\textwidth}
    \begin{tabular}{@{}>{\raggedright\arraybackslash}p{1.22in}
                    c
                    >{\centering\arraybackslash}p{1.12in}
                    >{\centering\arraybackslash}p{1.12in}
                    >{\centering\arraybackslash}p{1.12in}
                    >{\centering\arraybackslash}p{1.12in}
                    >{\centering\arraybackslash}p{1.12in}@{}}
        \toprule
        \multirow{2}{*}{Backbone}
        & \multirow{2}{*}{Metric}
        & \multicolumn{3}{c}{Segmentation}
        & Registration
        & 3D VISTA (MONAI) \\
        \cmidrule(lr){3-5}\cmidrule(lr){6-6}\cmidrule(l){7-7}
        & & Longitudinal MRI & Brain MRI & Lung PET/CT
          & Longitudinal MRI & Lung PET/CT \\
        \midrule

        \rowcolor{ToolHeadBg}
        \multicolumn{7}{@{}l}{\textbf{Gemini 3 Flash Preview}} \\
        & Task
            & \tcdown{0.28}{0.02}
            & \tcdown{0.65}{0.02}
            & \tcup{0.47}{0.02}
            & \tcdown{0.28}{0.05}
            & \tcup{0.50}{0.05} \\
        & Strict
            & \tcdown{0.05}{0.02}
            & \tcdown{0.56}{0.03}
            & \tcdown{0.09}{0.04}
            & \tcflat{0.07}
            & \tcflat{0.13} \\
        & Loc.
            & \tcna
            & \tcdown{0.85}{0.02}
            & \tcdown{0.24}{0.02}
            & \tcna
            & \tcup{0.50}{0.24} \\

        \addlinespace[0.25em]
        \rowcolor{ToolHeadBg}
        \multicolumn{7}{@{}l}{\textbf{Gemini 3.1 Pro Preview}} \\
        & Task
            & \tcdown{0.34}{0.03}
            & \tcup{0.79}{0.06}
            & \tcup{0.56}{0.06}
            & \tcdown{0.35}{0.01}
            & \tcup{0.61}{0.11} \\
        & Strict
            & \tcup{0.09}{0.01}
            & \tcdown{0.66}{0.07}
            & \tcup{0.20}{0.06}
            & \tcflat{0.06}
            & \tcup{0.22}{0.08} \\
        & Loc.
            & \tcna
            & \tcdown{0.87}{0.09}
            & \tcup{0.36}{0.15}
            & \tcna
            & \tcup{0.48}{0.27} \\

        \addlinespace[0.25em]
        \rowcolor{ToolHeadBg}
        \multicolumn{7}{@{}l}{\textbf{Qwen3.5-35B-A3B}} \\
        & Task
            & \tcdown{0.23}{0.10}
            & \tcup{0.65}{0.12}
            & \tcup{0.41}{0.06}
            & \tcup{0.30}{0.03}
            & \tcup{0.44}{0.09} \\
        & Strict
            & \tcdown{0.12}{0.06}
            & \tcup{0.21}{0.06}
            & \tcdown{0.00}{0.01}
            & \tcup{0.21}{0.08}
            & \tcflat{0.01} \\
        & Loc.
            & \tcna
            & \tcup{0.13}{0.02}
            & \tcdown{0.00}{0.01}
            & \tcna
            & \tcup{0.36}{0.35} \\

        \bottomrule
    \end{tabular}
    \end{adjustbox}
    \vspace{-3pt}
    }
    \caption{\textbf{Track B: Advanced Operations.}
    Each cell reports the score after enabling the specified advanced operation and its change relative to the Table~\ref{tab:main_results} viewer-native score for the same backbone, except Registration on Longitudinal MRI, which is compared with an additional unregistered LUMIERE control.
    Green, red, and gray cells indicate improvement, degradation, and no change, respectively.
    Task denotes final task accuracy, Strict denotes task-plus-evidence correctness, and Loc. denotes localization where available.}
    \vspace{-1pt}
    \label{tab:tooluse_trackb}
\end{table*}

\section{Failure Case Analysis}\label{sec:software_use_failure}

Beyond pre-selected image interpretation, medical imaging agents are expected to execute a clinically coherent software workflow: choose the right operation, apply it in the right image space, preserve the resulting state, and verify that the artifact supports the final answer. The incorrect advanced-operation episodes expose four recurring failure modes, plus a separate run-to-run instability pattern.

\noindent\textbf{Workflow intent management.} In a longitudinal MRI case (Appendix~\ref{app:reg_objective_drift}), the agent made plausible registration calls but appeared to drift from cross-timepoint response assessment into a mixture of longitudinal CT1 registration, CT1-derived FLAIR resampling, and intra-timepoint alignment. The operations were locally reasonable, but the composed workflow was no longer well aligned with the clinical question.

\noindent\textbf{Spatial and parametric action grounding.} In another longitudinal case (Appendix~\ref{app:pseudo_quantification}), the agent converted weak 2D screen boxes into RAS coordinates and then reported millimeter-scale lesion diameters and SPD change. This is not merely noisy measurement; it is invalid elevation of poorly grounded screen evidence into quantitative clinical evidence without validated boundaries, registration, or a measurement procedure.

\noindent\textbf{Stateful artifact management.} In a registered MRI fusion case (Appendix~\ref{app:transform_layer_misbinding}), transform objects, transformed nodes, viewer layers, and final rationale became inconsistent: a CT1 transform was requested, the executed trace resolved to a FLAIR transform, and the displayed fusion did not match the cited CT1 evidence. The failure is software-state misbinding, not only imperfect registration quality.

\noindent\textbf{Operation-result calibration and verification.} In a lung PET/CT case (Appendix~\ref{app:self_confirming_segmentation}), an early VISTA3D label-prompt segmentation was close to the reference lesion, but the agent discarded it and later used an incorrect manual point to prompt a self-confirming segmentation. The segmentation output was not uniformly bad; the agent failed to arbitrate between operation outputs and anatomical plausibility.

\noindent\textbf{Procedural stability of advanced operations.} Repeated runs of the same registration and segmentation cases show an additional failure mode: the agent does not reliably compose advanced operations into a stable workflow. In Appendix~\ref{app:tool_use_instability}, only 3 of 10 registration runs complete the expected BRAINSFit $\rightarrow$ resample/apply-transform $\rightarrow$ registered-fusion sequence, and segmentation runs vary from plausible masks to seed-only or no-mask outputs (Figures~\ref{fig:registration_workflow_inconsistency}--\ref{fig:segmentation_10_runs_comparison}).

Together, these cases show that medical VLM agents lack more than visual diagnostic skill. They lack reliable software-operation competence: objective preservation, spatial grounding, state tracking, operation-output calibration, and procedural control. These errors are hard to diagnose from final answers alone but are exposed by auditable traces of operation arguments, viewer states, generated artifacts, and evidence chains.

\section{Discussion and Conclusion}
\noindent\textbf{Discussion.} The introduction of \bench{} and \framework{} shifts the evaluation paradigm of medical vision-language models from isolated, static-image recognition to dynamic reasoning inside professional medical software. Our findings highlight several critical implications for the future of medical AI. 
\begin{itemize}[leftmargin=1.3em, topsep=2pt, itemsep=2pt]
    \item \textbf{Evidence-constrained evaluation exposes a stricter failure mode:} A slide-level or study-level answer can be correct while the supporting evidence is wrong, missing, or ungrounded. The Strict metrics make this distinction explicit by requiring the agent to provide checkable slices, coordinates, regions, or longitudinal evidence in addition to the task answer. This is the core reason the benchmark evaluates auditability rather than only label accuracy.
    \item \textbf{Advanced-operation access is not workflow competence:} Providing current VLMs with advanced segmentation operators, registration workflows, or MONAI-based expert-model operations does not automatically improve diagnostic accuracy. Some operations improve intermediate localization or evidence scores, but robust advanced-operation execution still requires fine-grained spatial grounding, correct parameterization, software-state tracking, and verification of returned artifacts.
    \item \textbf{Professional software operation is a distinct capability:} Operating 3D Slicer or QuPath is not equivalent to answering a question about a rendered screenshot. The agent must preserve the clinical objective while manipulating persistent viewer state, selecting regions, converting coordinates, and recording evidence. These behaviors are directly visible in the runtime trace.
    \item \textbf{Connecting evaluation to traceable interaction:} \bench{} evaluates agents in a full-study, software-mediated setting where actions, viewer state, and evidence artifacts are explicitly recorded. This interaction contract is closer to the substrate needed for human-in-the-loop imaging systems than pre-selected static 2D benchmarks, and can support exploratory applications such as \copilot{} in which agent actions and evidence remain inspectable by clinicians.
\end{itemize}

\vspace{-10pt}
\noindent\paragraph{Conclusion.} Medical agents must do more than answer localized questions about pre-selected images. They must actively operate professional medical software, navigate full volumetric studies or whole-slide images, gather comprehensive information, and leave an auditable trail of evidence for their decisions. \framework{} provides a secure, bounded and auditable runtime for this complex interaction, and \bench{} offers a rigorous, controlled environment to evaluate it. Our current results suggest that software-native full-study reasoning is already a tangible reality for frontier VLM agents, even as reliable, evidence-grounded and advanced-operation execution remains a compelling challenge for the community to solve.

\newpage

\bibliographystyle{unsrtnat}
\bibliography{radopenclaw_neurips}

\FloatBarrier

\appendix

\clearpage

\startcontents[appendices]

\section*{Appendix}

\printcontents[appendices]{}{1}{\setcounter{tocdepth}{2}}

\newpage
\section{Limitations and Roadmap} 
The benchmark is designed for controlled evaluation rather than clinical deployment. The pathology experiments currently emphasize viewer-native QuPath navigation and coordinate-grounded evidence acquisition; richer pathology-specific advanced operations and larger whole-slide cohorts remain important extensions. The runtime-agnostic track is useful for comparing alternative full-study pipelines, but such methods do not necessarily produce the same auditable software trace. The current scoring protocols also use public dataset labels and hidden annotations as reference standards, which are necessary for reproducibility but remain proxies for full clinical adjudication. Future extensions should broaden the modality coverage, increase pathology and longitudinal cohorts, add multi-turn clinician-agent interaction, and integrate additional vetted operations from the MONAI and medical imaging ecosystems.

\section{Runtime Interface and Safety Constraints}\label{app:runtime_interface}

To maintain a bounded and legible interface, the callable surface remains strictly explicit. Functions already supported by 3D Slicer are wrapped via documented WebServer REST endpoints~\citep{slicer_webserver_docs}, which process HTTP requests and responses to allow external control. Operations that are not cleanly covered by this REST interface, such as DICOM import, quantitative measurement, registration, segmentation, and DICOM SEG export, are exposed through named bridge handlers. The QuPath registry similarly exposes named operations such as loading a case or project, loading an image, setting image type, zooming, panning, focusing the viewer, capturing the current view, converting viewer points to whole-slide coordinates, recording the current view as evidence, sleeping, and closing QuPath. Crucially, this runtime is deliberately restrictive. While 3D Slicer includes an embedded Python console, allowing an agent to generate and execute arbitrary code would enlarge the attack surface, weaken auditability, and complicate deployment. Therefore, \framework{} exposes predefined operations only: it prohibits raw Python execution in the Slicer interface and does not expose direct Groovy or arbitrary-script execution in QuPath.

\section{Benchmark Modules, Dataset Sources, and Scoring}\label{app:benchmark_modules_scoring}
This appendix gives the public dataset sources and benchmark-specific scoring rules behind the task-family names used in the main text; Table~\ref{tab:module_summary} summarizes the benchmark modules, input units, case counts, primary tasks, and evidence requirements.

\begin{table*}[t]
    \centering
    {\scriptsize
    \setlength{\tabcolsep}{2.8pt}
    \renewcommand{\arraystretch}{1.16}
    \begin{adjustbox}{max width=\textwidth}
    \begin{tabular}{@{}p{0.68in} p{0.92in} p{1.12in} c p{1.36in} p{2.30in} p{1.55in}@{}}
        \toprule
        \rowcolor{AppBlockBg}
        Domain & Source & Input unit & \#Cases & Primary task & Evidence / Strict scoring & Localization scoring \\
        \midrule
        \rowcolor{AppDetailBg}
        Radiology & LUMIERE~\citep{suter2022lumiere} & BASELINE/FOLLOWUP brain MRI & 139 & RANO response category~\citep{wen2010rano} with lesion-state fields & Strict: RANO label is correct and all scorable FOLLOWUP-study evidence fields are correct: non-measurable lesion, T2 progression, current/new target lesion, and current/new non-target lesion. & Not required. \\
        Radiology & UCSF-PDGM~\citep{calabrese2022ucsfpdgm} & Multi-sequence brain MRI & 495 & Case-level tumor diagnosis & Strict: diagnosis is correct and KEY\_SLICE\_AXIAL\_MM, KEY\_SLICE\_SAGITTAL\_MM, and KEY\_SLICE\_CORONAL\_MM fall within the hidden tumor-mask RAS axis ranges. & Parsed RAS point lies inside the hidden tumor mask after RAS-to-voxel conversion. \\
        \rowcolor{AppDetailBg}
        Radiology & NSCLC~\citep{bakr2018nsclc} & Paired PET/CT study & 162 & Five structured labels: tumor location, pathological T stage, pathological N stage, histology, and grade & In NSCLC, the evidence block contains the same three key-slice offsets and RAS point. The evidence gate is stricter than slice-only evidence: all three key slices must be correct and the RAS point must fall inside the hidden lesion mask. The Strict score is not an all-or-nothing five-label exact match; if this evidence gate is satisfied, the case receives the five-question partial score \#correct/5, and otherwise it receives 0. & Parsed RAS point lies inside the hidden lesion mask after RAS-to-voxel conversion. \\
        Pathology & BRACS~\citep{brancati2022bracs} & Breast whole-slide image & 113 & Seven-class slide diagnosis: N, PB, UDH, FEA, ADH, DCIS, or IC & Strict: the slide-level BRACS label is correct and, for non-N evidence-scored cases, at least one returned WSI coordinate falls inside a QuPath annotation ROI whose normalized label matches the ground-truth class. For N cases, localizations must be empty or are not required, depending on the protocol. & On ground-truth positive slides, at least one output coordinate hits a correct ground-truth ROI. \\
        \rowcolor{AppDetailBg}
        Pathology & CAMELYON17~\citep{bandi2019camelyon17,litjens2018camelyon} & Lymph-node whole-slide image & 550 & Tumor presence and largest metastasis category & Strict: the tumor-presence and largest-metastasis outputs are correct and, for positive slides with a ground-truth evidence source, at least one returned coordinate hits the tumor mask or XML polygon and avoids exclusion regions. For negative slides, no tumor coordinate is required and localizations should be empty. & On ground-truth positive slides, at least one tumor coordinate hits the ground-truth tumor mask or polygon. \\
        \bottomrule
    \end{tabular}
    \end{adjustbox}
    }
    \caption{\textbf{Benchmark modules and scoring.} \bench{} evaluates agents that operate professional medical software and return both task answers and checkable evidence. The reported case counts correspond to the benchmark-eligible units used for evaluation. For NSCLC, the benchmark uses 162 eligible cases after excluding AMC cases by default; including AMC would yield 211 cases. For LUMIERE, the benchmark starts from 360 longitudinal questions and retains 139 after applying the evidence-availability and label-frequency filters used by the current evaluation protocol. For radiology, \textbf{Strict} denotes answer correctness under benchmark-specific evidence constraints. For pathology, evidence and localization are scored by deterministic point lookup against QuPath~\citep{bankhead2017qupath} annotations, masks, or XML polygons rather than by human or LLM judging.}
    \label{tab:module_summary}
\end{table*}
\FloatBarrier
\subsection{Module construction and evaluation protocols}

The benchmark spans radiology and pathology tasks evaluated under deterministic answer and evidence protocols. The brain MRI setting includes LUMIERE~\citep{suter2022lumiere} longitudinal MRI for lesion-evolution reasoning under RANO response criteria~\citep{wen2010rano} and UCSF-PDGM~\citep{calabrese2022ucsfpdgm}, a preoperative multi-sequence brain tumor MRI dataset, for case-level diagnosis over a fixed label set. The lung PET/CT setting uses the NSCLC radiogenomics dataset~\citep{bakr2018nsclc}, a paired non-small cell lung cancer PET/CT cohort with pathology annotations, for five structured prediction tasks: tumor location, pathological T stage, pathological N stage, histology, and histopathological grade. The pathology setting uses QuPath to evaluate whole-slide image navigation and evidence acquisition: BRACS~\citep{brancati2022bracs} requires seven-class slide-level diagnosis and one or more evidence coordinates for non-N cases, while CAMELYON17~\citep{bandi2019camelyon17,litjens2018camelyon} requires slide-level tumor presence, largest metastasis category, and tumor coordinates for positive predictions. All experiments in this paper use the MCQ or structured-answer protocol, which provides a canonical output schema and enables deterministic scoring.

For radiology, the agent is required to output both the final task answer and an evidence block. \textbf{Strict} is the evidence-constrained score: a case is credited only when the final task answer and the required evidence are both correct, with benchmark-specific definitions. In UCSF-PDGM, the evidence block contains KEY\_SLICE\_AXIAL\_MM, KEY\_SLICE\_SAGITTAL\_MM, KEY\_SLICE\_CORONAL\_MM, and RAS: [R,A,S]. Strict requires the diagnostic class to be correct and the three key-slice offsets to fall within the hidden tumor-mask RAS axis ranges; the RAS point is not included in Strict and is scored separately as localization. In NSCLC, the evidence block contains the same three key-slice offsets and RAS point, but the evidence condition is stricter: all three key slices must be correct and the RAS point must fall inside the hidden lesion mask. The Strict score is the five-question partial score \#correct/5 only if this exact evidence condition is satisfied; otherwise the case receives 0. In LUMIERE, there is no key-slice or RAS evidence block. Instead, evidence consists of CURRENT-study YES/NO fields for non-measurable lesion, T2 progression, current target lesion, new target lesion, current non-target lesion, and new non-target lesion. Strict requires the RANO category to be correct and all scorable evidence fields to be correct; localization is not reported for LUMIERE.

For pathology, the agent outputs a slide-level answer and, when relevant, one or more whole-slide evidence coordinates. BRACS scores evidence on evidence-scored cases by checking whether an output coordinate falls inside a QuPath annotation ROI whose normalized label matches the ground-truth class; localization is computed on ground-truth positive slides by whether at least one coordinate hits the correct ground-truth ROI. CAMELYON17 scores evidence by checking whether the output coordinate falls inside a ground-truth tumor mask or XML polygon and outside exclusion regions; localization is computed on ground-truth positive slides by the same hit criterion. These pathology evidence scores are deterministic and do not use human review or LLM judging.

\subsection{2D slice-montage input for Track C}\label{app:slice_montage}

The 2D slice-montage baseline is the simplest runtime-agnostic alternative to interactive viewer control. It compresses a volumetric study into a static grid of representative slices, allowing a general VLM to inspect the case without calling 3D Slicer~\citep{fedorov2012slicer}. This format is useful for testing whether visible slice-level cues survive compression, but it deliberately removes important elements of clinical workflow: dynamic scrolling, cross-plane confirmation, series switching, fusion state, measurements, coordinate conversion, and replayable evidence acquisition.

\begin{figure}
    \centering
    \includegraphics[width=0.5\linewidth]{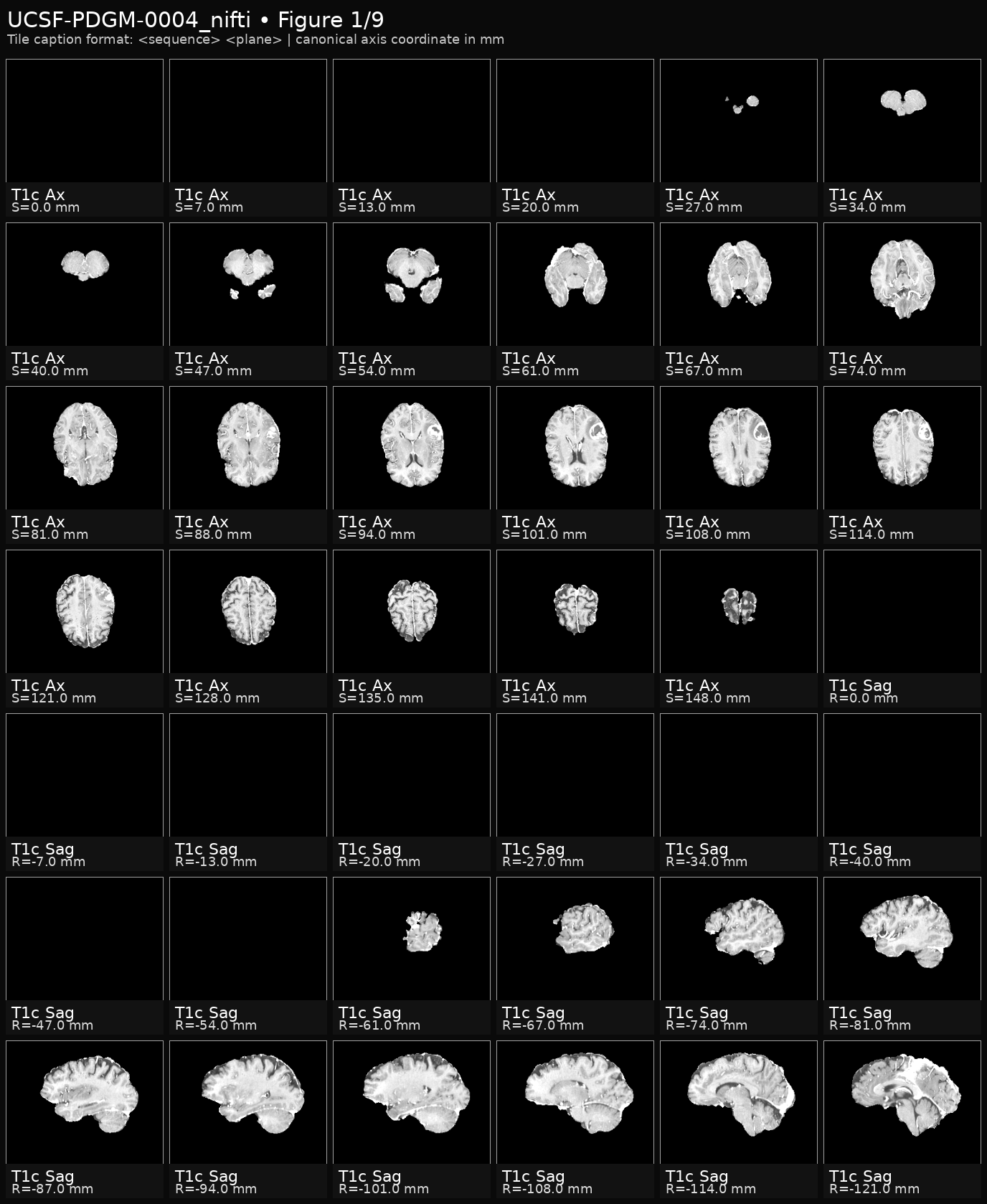}
    \caption{\textbf{2D slice-montage baseline for Track C.} A volumetric MRI or CT study is rendered as a static tiled overview of representative slices, analogous to a compact film sheet. The baseline lets non-viewer methods consume the raw case through a single image input, but it does not preserve interactive viewer state, key-slice navigation, coordinate-grounded evidence, or longitudinal/fusion operations.}
    \label{fig:slice_montage}
\end{figure}

\begin{table*}[t]
    \centering
    {\tiny
    \setlength{\tabcolsep}{1.7pt}
    \renewcommand{\arraystretch}{1.06}
    \begin{adjustbox}{max width=\textwidth}
    \begin{tabular}{@{}>{\raggedright\arraybackslash}p{1.35in}
                    DDDDDD c S
                    c S G
                    c DDDDD S G@{}}
        \toprule
        \multirow{3}{*}{Model}
            & \multicolumn{8}{c}{Longitudinal MRI}
            & \multicolumn{3}{c}{Brain MRI}
            & \multicolumn{8}{c}{Lung PET/CT} \\
        \cmidrule(lr){2-9}\cmidrule(lr){10-12}\cmidrule(l){13-20}
            & \multicolumn{6}{c}{Evidence fields}
            & \multicolumn{2}{c}{Answer + evidence}
            & \multicolumn{2}{c}{Answer + evidence}
            & \multicolumn{1}{c}{Grounding}
            & \multicolumn{6}{c}{Answer + subtasks}
            & \multicolumn{2}{c}{Evidence + grounding} \\
        \cmidrule(lr){2-7}\cmidrule(lr){8-9}
        \cmidrule(lr){10-11}\cmidrule(lr){12-12}
        \cmidrule(lr){13-18}\cmidrule(l){19-20}
            & \makecell{NML}
            & \makecell{T2\\Prog.}
            & \makecell{Cur.\\TL}
            & \makecell{New\\TL}
            & \makecell{Cur.\\NTL}
            & \makecell{New\\NTL}
            & Task
            & Strict
            & Task
            & Strict
            & Loc.
            & Task
            & \makecell{Tumor\\Loc.}
            & \makecell{T\\Stage}
            & \makecell{Path.\\N}
            & Hist.
            & Grade
            & Strict
            & Loc. \\
        \midrule

        \rowcolor{AppBlockBg}
        \multicolumn{20}{@{}l}{\textbf{Closed-source models}} \\
        GPT-5.4 mini
            & \textbf{0.58} & 0.20 & \textbf{0.86} & 0.00 & 0.43 & 0.00 & 0.23 & 0.05
            & 0.38 & 0.18 & 0.37
            & 0.31 & 0.23 & 0.32 & 0.42 & 0.37 & 0.20 & 0.00 & 0.00 \\
        GPT-5.5
            & 0.22 & 0.39 & 0.73 & 0.25 & \textbf{0.95} & 0.02 & 0.33 & 0.04
            & 0.71 & 0.66 & 0.91
            & \textbf{0.55} & \textbf{0.50} & \textbf{0.50} & \textbf{0.72} & 0.62 & 0.43 & 0.03 & 0.07 \\
        Gemini 3 Flash Preview
            & 0.16 & 0.61 & 0.71 & \textbf{0.40} & 0.88 & \textbf{0.30} & 0.30 & 0.07
            & 0.67 & 0.59 & 0.87
            & 0.45 & 0.39 & 0.25 & 0.46 & 0.69 & \textbf{0.50} & 0.13 & \textbf{0.26} \\
        Gemini 3.1 Pro Preview
            & 0.21 & 0.46 & 0.85 & 0.24 & 0.72 & 0.10 & \textbf{0.37} & 0.08
            & \textbf{0.73} & \textbf{0.73} & \textbf{0.96}
            & 0.50 & \textbf{0.50} & 0.36 & 0.62 & 0.68 & 0.35 & \textbf{0.14} & 0.21 \\
        Claude Sonnet 4.6
            & \textbf{0.58} & 0.33 & 0.56 & 0.16 & 0.73 & 0.17 & 0.33 & 0.17
            & 0.38 & 0.18 & 0.36
            & 0.26 & 0.18 & 0.09 & 0.16 & 0.46 & 0.36 & 0.00 & 0.00 \\
        Claude Opus 4.7
            & 0.53 & \textbf{0.75} & 0.67 & 0.14 & 0.80 & 0.13 & 0.28 & \textbf{0.18}
            & 0.37 & 0.11 & 0.68
            & 0.43 & 0.38 & 0.22 & 0.40 & \textbf{0.73} & 0.38 & 0.01 & 0.01 \\
        \rowcolor{AppSummaryBg}
        \emph{Mean over reported}
            & 0.38 & 0.46 & 0.73 & 0.20 & 0.75 & 0.12 & 0.31 & 0.10
            & 0.54 & 0.41 & 0.69
            & 0.42 & 0.36 & 0.29 & 0.46 & 0.59 & 0.37 & 0.05 & 0.09 \\

        \addlinespace[0.2em]
        \rowcolor{AppBlockBg}
        \multicolumn{20}{@{}l}{\textbf{Open-weight / open-source models}} \\
        Qwen3.5-35B-A3B
            & \textbf{0.88} & 0.23 & 0.57 & 0.00 & 0.00 & 0.00 & \textbf{0.33} & \textbf{0.18}
            & \textbf{0.53} & 0.15 & 0.11
            & 0.35 & 0.22 & \textbf{0.25} & 0.46 & 0.54 & 0.30 & \textbf{0.01} & \textbf{0.01} \\
        Qwen3.5-27B
            & 0.84 & \textbf{0.46} & 0.76 & 0.00 & 0.11 & 0.00 & 0.21 & 0.09
            & \textbf{0.53} & \textbf{0.27} & \textbf{0.36}
            & \textbf{0.37} & 0.22 & 0.22 & \textbf{0.54} & 0.54 & \textbf{0.33} & 0.00 & 0.00 \\
        Qwen3.5-9B
            & 0.71 & 0.23 & 0.65 & 0.13 & 0.26 & 0.08 & 0.17 & 0.08
            & 0.35 & 0.05 & 0.06
            & 0.16 & 0.08 & 0.17 & 0.22 & 0.21 & 0.13 & \textbf{0.01} & \textbf{0.01} \\
        Qwen3.5-4B
            & 0.72 & 0.29 & 0.86 & \textbf{0.23} & \textbf{0.28} & \textbf{0.16} & 0.22 & 0.11
            & 0.32 & 0.06 & 0.03
            & 0.20 & 0.17 & 0.10 & 0.21 & 0.33 & 0.19 & 0.00 & 0.00 \\
        MedGemma 1.5 4B IT
            & 0.64 & 0.45 & 0.75 & 0.06 & 0.17 & 0.02 & 0.16 & 0.02
            & 0.45 & 0.09 & 0.00
            & 0.16 & 0.11 & 0.05 & 0.17 & 0.27 & 0.20 & 0.00 & 0.00 \\
        MedGemma 27B IT
            & 0.75 & 0.00 & 0.96 & 0.00 & 0.10 & 0.02 & 0.11 & 0.00
            & 0.30 & 0.00 & 0.03
            & 0.16 & \textbf{0.28} & 0.13 & 0.27 & 0.10 & 0.02 & 0.00 & 0.00 \\
        Lingshu-32B
            & 0.82 & 0.02 & \textbf{0.98} & 0.00 & 0.02 & 0.00 & 0.16 & 0.00
            & 0.36 & 0.08 & 0.02
            & 0.31 & 0.15 & 0.20 & 0.38 & \textbf{0.57} & 0.23 & 0.00 & 0.00 \\
        \rowcolor{AppSummaryBg}
        \emph{Mean over reported}
            & 0.76 & 0.24 & 0.79 & 0.06 & 0.13 & 0.04 & 0.19 & 0.07
            & 0.41 & 0.10 & 0.09
            & 0.24 & 0.18 & 0.16 & 0.32 & 0.37 & 0.20 & 0.00 & 0.00 \\

        \bottomrule
    \end{tabular}
    \end{adjustbox}
    }
    \caption{\textbf{Detailed viewer-native radiology metrics.}
    This appendix table expands Table~\ref{tab:main_results} with lesion-state evidence fields for Longitudinal MRI and subtask-level metrics for Lung PET/CT.
    NML denotes non-measurable-lesion evidence-field accuracy.
    Gray columns are fine-grained evidence or subtask metrics, yellow columns are \textbf{Strict} evidence-constrained scores, and blue columns are localization metrics.
    Bold numbers indicate the best reported score within each model group and metric column; summary rows report means over available entries. Model groups in this appendix are organized by access type rather than the scope grouping used in Table~\ref{tab:main_results}.}
    \label{tab:main_results_detailed}
\end{table*}

\section{Runtime-Agnostic Track Results}\label{app:runtime_agnostic_track}

Table~\ref{tab:open_method} reports the full Track C comparison between 3D Slicer viewer-native runs and runtime-agnostic alternatives under the same answer and evidence schema. For runtime-agnostic baselines, the model returns the same canonical answer and evidence fields directly from the supplied rendered or volumetric input; these runs are scored by the same parsers but do not produce a replayable viewer-state trace. This appendix table is a diagnostic companion to the main viewer-native and advanced-operation results: the differences are task-dependent, and the table mainly documents coverage of pipelines that do not use \framework{} rather than introducing a separate main claim.

\begin{table*}[t]
    \centering
    {\scriptsize
    \setlength{\tabcolsep}{2.0pt}
    \renewcommand{\arraystretch}{1.10}
    \begin{adjustbox}{max width=\textwidth}
    \begin{tabular}{@{}l@{\hspace{0.55em}}cc ccc ccc@{}}
        \toprule
        \multirow{2}{*}{Backbone}
            & \multicolumn{2}{c}{Longitudinal MRI}
            & \multicolumn{3}{c}{Brain MRI}
            & \multicolumn{3}{c}{Lung PET/CT} \\
        \cmidrule(lr){2-3}\cmidrule(lr){4-6}\cmidrule(l){7-9}
            & Task & Strict
            & Task & Strict & Loc.
            & Task & Strict & Loc. \\
        \midrule

        \rowcolor{OpenBlockBg}
        \multicolumn{9}{@{}l}{\textbf{2D slice-montage}} \\

        \rowcolor{OpenSubBg}
        \multicolumn{9}{@{}l}{\emph{General-purpose backbones}} \\
        GPT-5.4 mini
            & \tcdown{0.17}{0.06} & \tcdown{0.00}{0.05}
            & \tcup{0.41}{0.03} & \tcup{0.25}{0.07} & \tcdown{0.00}{0.37}
            & \tcup{0.40}{0.09} & \tcup{0.01}{0.01} & \tcflat{0.00} \\
        GPT-5.5
            & \tcdown{0.01}{0.32} & \tcdown{0.00}{0.04}
            & \tcup{0.84}{0.13} & \tcup{0.79}{0.13} & \tcup{0.97}{0.06}
            & \tcdown{0.54}{0.01} & \tcup{0.13}{0.10} & \tcup{0.23}{0.16} \\
        Gemini 3.1 Pro Preview
            & \tcdown{0.22}{0.15} & \tcdown{0.00}{0.08}
            & \tcup{0.76}{0.03} & \tcdown{0.71}{0.02} & \tcdown{0.81}{0.15}
            & \tcdown{0.49}{0.01} & \tcdown{0.11}{0.03} & \tcdown{0.13}{0.08} \\
        Gemini 3 Flash Preview
            & \tcdown{0.13}{0.17} & \tcdown{0.00}{0.07}
            & \tcdown{0.52}{0.15} & \tcdown{0.43}{0.16} & \tcdown{0.77}{0.10}
            & \tcup{0.51}{0.06} & \tcdown{0.07}{0.06} & \tcdown{0.08}{0.18} \\
        Claude Sonnet 4.6
            & \tcdown{0.12}{0.21} & \tcdown{0.00}{0.17}
            & \tcup{0.52}{0.14} & \tcup{0.19}{0.01} & \tcup{0.38}{0.02}
            & \tcup{0.45}{0.19} & \tcup{0.06}{0.06} & \tcup{0.02}{0.02} \\
        Claude Opus 4.7
            & \tcdown{0.20}{0.08} & \tcdown{0.03}{0.15}
            & \tcup{0.58}{0.21} & \tcdown{0.08}{0.03} & \tcdown{0.05}{0.63}
            & \tcup{0.47}{0.04} & \tcflat{0.01} & \tcdown{0.00}{0.01} \\
        Qwen3.5-35B-A3B
            & \tcdown{0.19}{0.14} & \tcdown{0.12}{0.06}
            & \tcup{0.56}{0.03} & \tcup{0.38}{0.23} & \tcup{0.54}{0.43}
            & \tcdown{0.25}{0.10} & \tcup{0.02}{0.01} & \tcdown{0.00}{0.01} \\

        \addlinespace[0.15em]
        \rowcolor{OpenSubBg}
        \multicolumn{9}{@{}l}{\emph{Medical-domain backbones}} \\
        MedGemma 1.5 4B IT
            & \tcdown{0.12}{0.04} & \tcup{0.05}{0.03}
            & \tcdown{0.35}{0.10} & \tcdown{0.00}{0.09} & \tcflat{0.00}
            & \tcdown{0.09}{0.07} & \tcflat{0.00} & \tcflat{0.00} \\
        MedGemma 27B IT
            & \tcup{0.29}{0.18} & \tcflat{0.00}
            & \tcup{0.35}{0.05} & \tcflat{0.00} & \tcdown{0.00}{0.03}
            & \tcdown{0.11}{0.05} & \tcflat{0.00} & \tcflat{0.00} \\
        Lingshu-32B
            & \tcup{0.24}{0.08} & \tcup{0.02}{0.02}
            & \tcup{0.38}{0.02} & \tcdown{0.06}{0.02} & \tcdown{0.00}{0.02}
            & \tcup{0.34}{0.03} & \tcflat{0.00} & \tcflat{0.00} \\

        \midrule
        \rowcolor{OpenBlockBg}
        \multicolumn{9}{@{}l}{\textbf{Native 3D VLM, Alt. only}} \\
        M3D
            & \oalt{0.21} & \oalt{0.10}
            & \oalt{0.40} & \oalt{0.10} & \oalt{0.36}
            & \oalt{0.18} & \oalt{0.07} & \oalt{0.33} \\
        RadFM
            & \oalt{0.25} & \oalt{0.00}
            & \oalt{0.51} & \oalt{0.10} & \oalt{0.36}
            & \oalt{0.25} & \oalt{0.00} & \oalt{0.00} \\

        \bottomrule
    \end{tabular}
    \end{adjustbox}
    }
    \caption{\textbf{Runtime-agnostic track.}
    Each paired cell reports the alternative runtime-agnostic score and its change relative to the matched 3D Slicer viewer-native run for the same backbone.
    Green, red, and gray cells indicate improvement, degradation, and no change, respectively.
    Task denotes final task accuracy, Strict denotes task-plus-evidence correctness, and Loc. denotes localization where available.
    The 2D slice-montage baseline renders volumetric studies as static tiled overview figures, while native 3D VLM rows report Alt.-only volumetric baselines without matched same-backbone Slicer controls.}
    \label{tab:open_method}
\end{table*}

\FloatBarrier
\section{Qualitative Workflow Illustrations}\label{app:workflow_illustrations}

\begin{figure}[!htbp]
    \centering
    \includegraphics[width=1\linewidth]{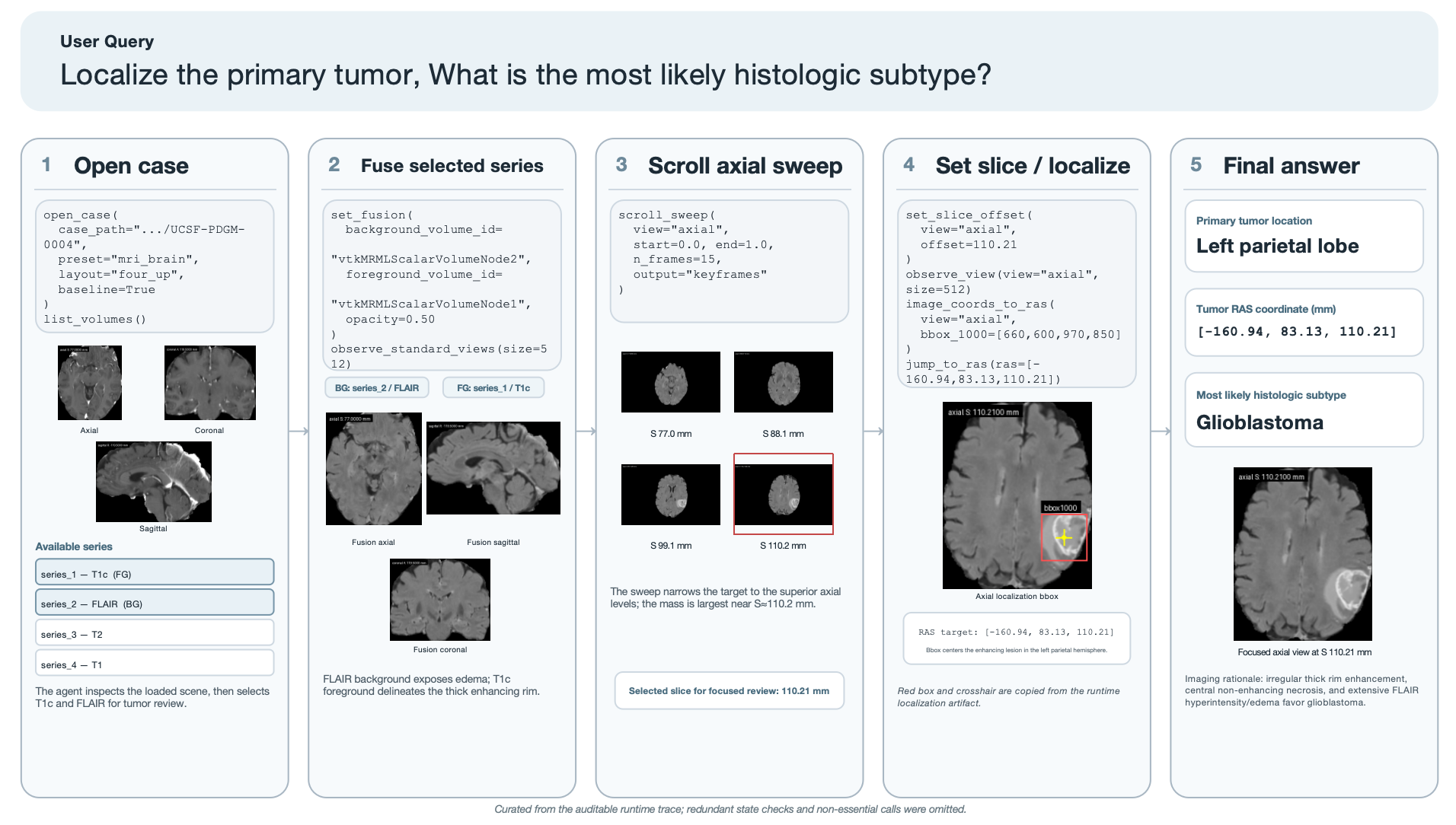}
    \caption{\textbf{Viewer-native Brain MRI example in UCSF-PDGM~\citep{calabrese2022ucsfpdgm}.} The agent opens the full multi-sequence case, selects T1c and FLAIR, uses fusion and axial scrolling to localize the dominant enhancing lesion, converts the selected image point to a RAS coordinate, and returns both tumor location and likely histologic subtype. No segmentation, registration, or MONAI expert-model operation is used; the example demonstrates that primitive viewer operations can support a replayable localization-and-diagnosis trace when the agent maintains the correct series and slice state.}
    \label{fig:viewer_native_brain_mri_example}
\end{figure}

\begin{figure}[!htbp]
    \centering
    \includegraphics[width=1\linewidth]{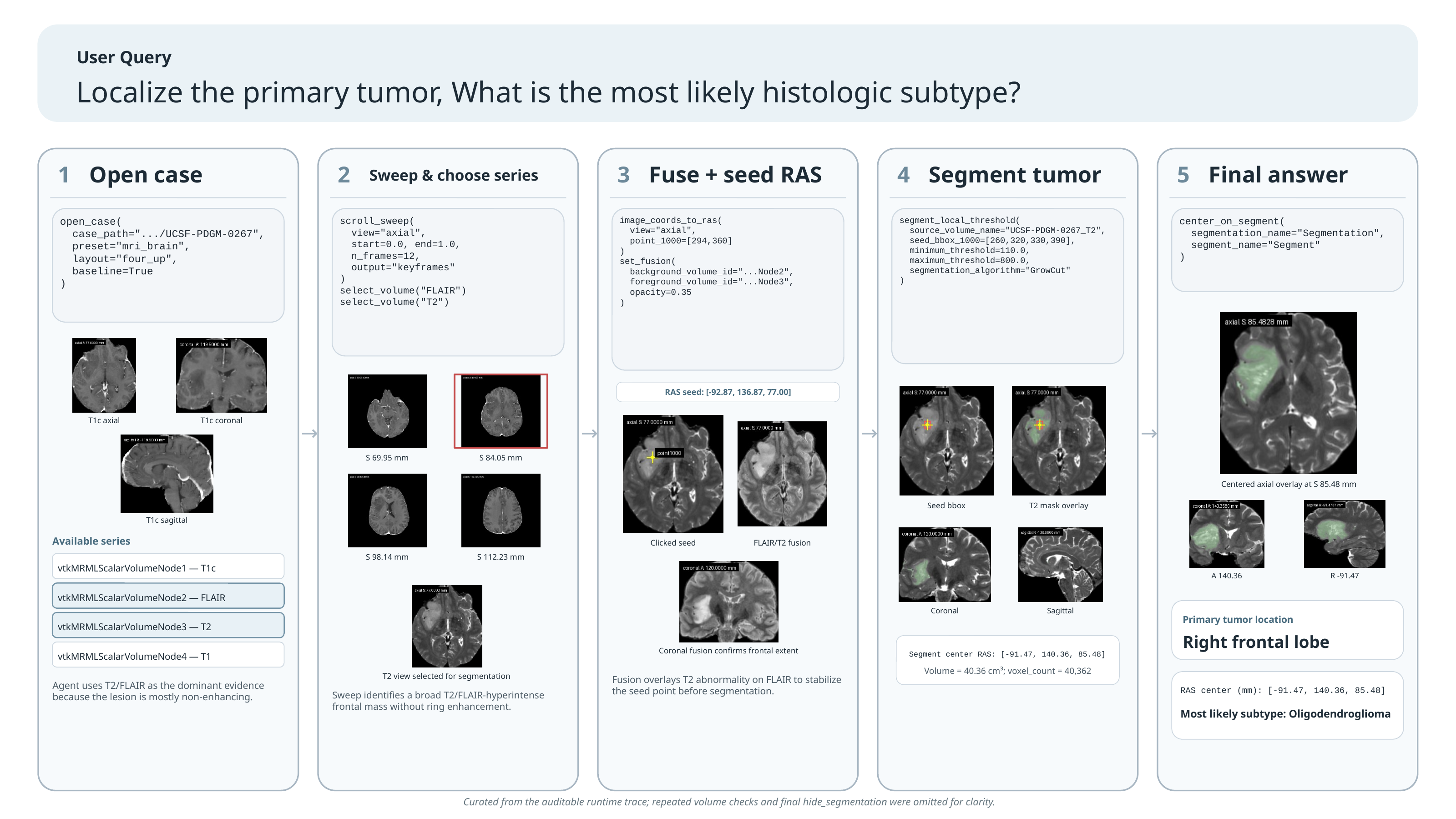}
    \caption{\textbf{Segmentation-assisted Brain MRI example.} After inspecting the UCSF-PDGM study and selecting FLAIR/T2-dominant evidence, the agent uses the fused viewer state to seed a RAS point and invokes the segmentation operation to generate a tumor mask. The final answer is grounded by the segment center, orthogonal views, and a reported RAS coordinate. The example illustrates how an advanced operation can convert an inspected lesion candidate into structured evidence, while still requiring the agent to choose the correct source volume, seed location, and interpretation.}
    \label{fig:tool_segmentation_brain_mri_example}
\end{figure}

\begin{figure}[!htbp]
    \centering
    \includegraphics[width=1\linewidth]{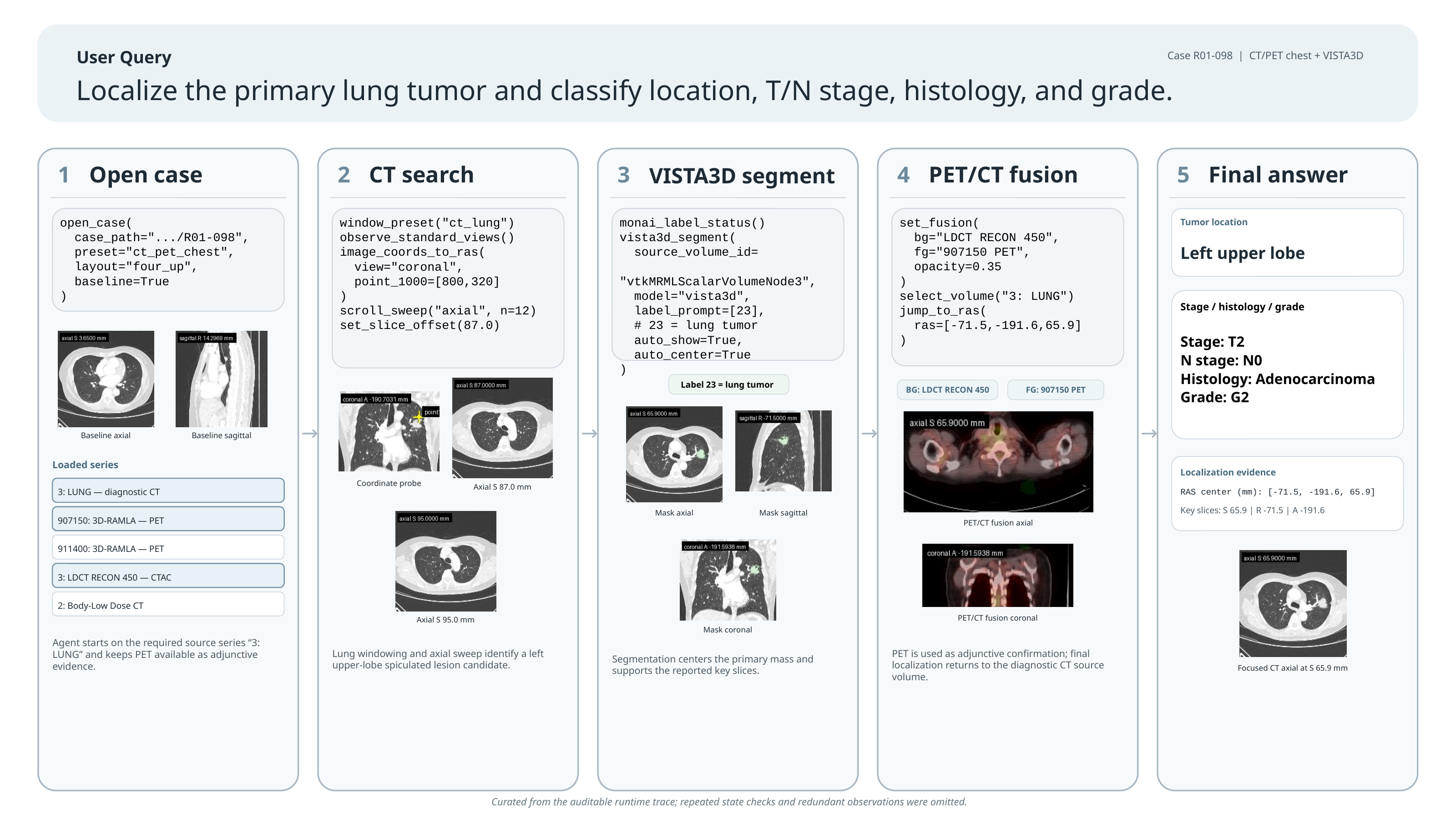}
    \caption{\textbf{MONAI 3D VISTA~\citep{cardoso2022monai,he2025vista3d} advanced-operation example for Lung PET/CT.} The agent first searches the diagnostic CT volume, invokes the MONAI 3D VISTA workflow with the lung-tumor label to obtain a lesion-centered segmentation, then checks the candidate against PET/CT fusion before returning tumor location, T/N stage, histology, grade, key slices, and RAS evidence. The figure illustrates the intended role of 3D VISTA in \framework{}: it is an auditable expert-model operation inside the Slicer workflow, not a replacement for source-series selection, fusion verification, or final clinical reasoning.}
    \label{fig:tool_vista_lung_example}
\end{figure}

This section provides concrete examples of the two interaction regimes used throughout the benchmark. Figure~\ref{fig:viewer_native_brain_mri_example} shows a viewer-native Brain MRI episode that uses only primitive 3D Slicer actions, while Figures~\ref{fig:tool_segmentation_brain_mri_example} and~\ref{fig:tool_vista_lung_example} illustrate advanced-operation workflows, including MONAI VISTA3D-based segmentation. These figures are not additional results; they clarify what an auditable episode looks like, what evidence is recorded, and where advanced operations enter the runtime.
\FloatBarrier

\FloatBarrier
\section{Operation-Budget Study}\label{app:tool_budget}
The following diagnostic study varies the maximum number of explicit operation calls available to the agent. Table~\ref{tab:tool_budget} reports this budget control.

\begin{table*}[!htbp]
    \centering
    {\scriptsize
    \setlength{\tabcolsep}{2.0pt}
    \renewcommand{\arraystretch}{1.08}
    \begin{adjustbox}{max width=\textwidth}
    \begin{tabular}{>{\raggedright\arraybackslash}p{1.55in} c DDDDDD c S c S G c DDDDD S G}
        \toprule
        Backbone & \makecell{Max\\ops.}
            & \multicolumn{8}{c}{\makecell{Longitudinal\\MRI}}
            & \multicolumn{3}{c}{\makecell{Brain\\MRI}}
            & \multicolumn{8}{c}{\makecell{Lung\\PET/CT}} \\
        \cmidrule(lr){3-10}\cmidrule(lr){11-13}\cmidrule(lr){14-21}
            & & \makecell{NML} & \makecell{T2\\Prog.} & \makecell{Cur.\\TL} & \makecell{New\\TL} & \makecell{Cur.\\NTL} & \makecell{New\\NTL} & Task & Strict
            & Task & Strict & Loc.
            & Task & \makecell{Tumor\\Loc.} & \makecell{T\\Stage} & \makecell{Path.\\N} & Hist. & Grade & Strict & Loc. \\
        \midrule
        \rowcolor{AppBlockBg}
        \multicolumn{21}{@{}l}{\textbf{Qwen3.5-35B-A3B}} \\
        \midrule
        Qwen3.5-35B-A3B & 1
            & 0.88 & 0.15 & 0.65 & 0.00 & 0.14 & 0.00 & 0.21 & 0.13
            & 0.41 & 0.11 & 0.00
            & 0.32 & 0.15 & 0.22 & 0.43 & 0.47 & 0.35 & 0.00 & 0.00 \\
        Qwen3.5-35B-A3B & 3
            & 0.82 & 0.21 & 0.55 & 0.00 & 0.11 & 0.04 & 0.21 & 0.10
            & 0.50 & 0.11 & 0.13
            & 0.30 & 0.17 & 0.25 & 0.47 & 0.42 & 0.20 & 0.00 & 0.00 \\
        Qwen3.5-35B-A3B & 7
            & 0.86 & 0.42 & 0.65 & 0.13 & 0.14 & 0.04 & 0.27 & 0.13
            & 0.50 & 0.15 & 0.13
            & 0.28 & 0.18 & 0.18 & 0.33 & 0.40 & 0.30 & 0.00 & 0.00 \\
        Qwen3.5-35B-A3B & 9
            & 0.86 & 0.48 & 0.74 & 0.38 & 0.14 & 0.00 & 0.25 & 0.17
            & 0.52 & 0.11 & 0.23
            & 0.36 & 0.25 & 0.30 & 0.50 & 0.47 & 0.28 & 0.00 & 0.02 \\
        Qwen3.5-35B-A3B & 10
            & 0.83 & 0.30 & 0.74 & 0.25 & 0.14 & 0.00 & 0.26 & 0.16
            & 0.46 & 0.13 & 0.25
            & 0.32 & 0.23 & 0.20 & 0.48 & 0.42 & 0.25 & 0.00 & 0.02 \\
        Qwen3.5-35B-A3B & 15
            & 0.87 & 0.36 & 0.61 & 0.13 & 0.14 & 0.04 & 0.24 & 0.12
            & 0.52 & 0.16 & 0.23
            & 0.37 & 0.23 & 0.30 & 0.62 & 0.42 & 0.28 & 0.00 & 0.00 \\
        Qwen3.5-35B-A3B & 20
            & 0.87 & 0.45 & 0.74 & 0.00 & 0.14 & 0.00 & 0.24 & 0.16
            & 0.46 & 0.11 & 0.17
            & 0.31 & 0.12 & 0.27 & 0.48 & 0.43 & 0.25 & 0.00 & 0.00 \\
        \midrule
        \rowcolor{AppBlockBg}
        \multicolumn{21}{@{}l}{\textbf{Gemini 3.1 Pro Preview}} \\
        \midrule
        Gemini 3.1 Pro Preview & 1
            & 0.44 & 0.39 & 0.87 & 0.13 & 0.41 & 0.04 & 0.23 & 0.13
            & 0.76 & 0.35 & 0.47
            & 0.55 & 0.57 & 0.27 & 0.75 & 0.65 & 0.52 & 0.00 & 0.02 \\
        Gemini 3.1 Pro Preview & 5
            & 0.86 & 0.06 & 0.87 & 0.00 & 0.14 & 0.04 & 0.18 & 0.10
            & 0.42 & 0.14 & 0.64
            & 0.56 & 0.53 & 0.27 & 0.78 & 0.70 & 0.50 & 0.08 & 0.14 \\
        Gemini 3.1 Pro Preview & 10
            & 0.57 & 0.21 & 0.84 & 0.38 & 0.30 & 0.04 & 0.29 & 0.13
            & 0.71 & 0.60 & 0.93
            & 0.54 & 0.58 & 0.33 & 0.72 & 0.65 & 0.42 & 0.11 & 0.28 \\
        Gemini 3.1 Pro Preview & 15
            & 0.70 & 0.12 & 0.74 & 0.38 & 0.22 & 0.04 & 0.27 & 0.07
            & 0.75 & 0.71 & 0.91
            & 0.52 & 0.57 & 0.30 & 0.68 & 0.70 & 0.33 & 0.10 & 0.30 \\
        \bottomrule
    \end{tabular}
    \end{adjustbox}
    }
    \caption{\textbf{Operation-budget study.} Only the explicit budget-control blocks for Qwen3.5-35B-A3B and Gemini 3.1 Pro Preview are shown. NML denotes non-measurable-lesion evidence-field accuracy.}
    \label{tab:tool_budget}
\end{table*}

\FloatBarrier
\section{Additional Failure Cases}\label{app:failure_cases}

The following cases expand the qualitative failure analysis in Section~\ref{sec:software_use_failure}. Each case is presented as a compact auditable case study: a case card states the clinical objective and observed failure signal, a trace excerpt shows the minimal logged software calls needed to diagnose the error, and a mechanism box links the trace to the missing agent capability.

\Needspace{14\baselineskip}
\subsection{Case F1: Registration objective drift in longitudinal MRI}\label{app:reg_objective_drift}

\begin{casecard}{F1 summary: workflow intent management}
\small
\begin{tabularx}{\linewidth}{@{}>{\bfseries}p{0.20\linewidth}X@{}}
Clinical objective & Longitudinal response assessment between week-015 and week-024 for Patient-009. \\
Ground truth / prediction & RANO complete response (CR) / predicted partial response (PR). \\
Trace signal & The agent created a plausible cross-timepoint CT1 transform, but then mixed longitudinal CT1 registration with CT1-derived FLAIR resampling and intra-timepoint FLAIR-to-CT1 alignment. \\
Why it matters & The final workflow no longer supported the intended longitudinal comparison, despite containing locally plausible registration calls. \\
\end{tabularx}
\end{casecard}

\begin{tracecompact}{Minimal auditable trace}
# Intended longitudinal CT1 registration.
brains_fit:
  moving = FOLLOWUP__week-024__CT1
  fixed  = BASELINE__week-015__CT1
  output = FOLLOWUP_to_BASELINE_CT1_transform

brains_resample:
  input     = FOLLOWUP__week-024__CT1
  reference = BASELINE__week-015__CT1
  transform = FOLLOWUP_to_BASELINE_CT1_transform
  output    = FOLLOWUP__week-024__CT1__aligned

# Drift: the CT1-derived longitudinal transform is applied to FLAIR.
brains_resample:
  input     = FOLLOWUP__week-024__FLAIR
  reference = BASELINE__week-015__CT1
  transform = FOLLOWUP_to_BASELINE_CT1_transform
  output    = FOLLOWUP__week-024__FLAIR__aligned

# Additional intra-timepoint registration changes the comparison objective.
brains_fit:
  moving = BASELINE__week-015__FLAIR
  fixed  = BASELINE__week-015__CT1
  output = BASELINE_FLAIR_to_CT1_transform

brains_resample:
  input     = BASELINE__week-015__FLAIR
  reference = BASELINE__week-015__CT1
  transform = BASELINE_FLAIR_to_CT1_transform
  output    = BASELINE__week-015__FLAIR__aligned

# Final displayed comparison is between two derived FLAIR volumes.
set_fusion:
  background = BASELINE__week-015__FLAIR__aligned
  foreground = FOLLOWUP__week-024__FLAIR__aligned
  opacity    = 0.5
\end{tracecompact}

\begin{mechanismbox}
\small
\textbf{Mechanism.} The failure is not the absence of registration calls. The agent repeatedly invoked registration operations and produced plausible intermediate artifacts, but the workflow objective became unstable. The clinical task required a longitudinal response assessment across timepoints, while the executed chain mixed longitudinal CT1 registration, CT1-derived FLAIR resampling, and intra-timepoint BASELINE FLAIR-to-CT1 alignment.

\textbf{Benchmark implication.} This exposes a workflow-intent failure: the agent can execute locally plausible software operations without preserving the semantic purpose of the clinical workflow. The auditable trace reveals the mismatch between the claimed comparison and the actual composed registration chain.
\end{mechanismbox}

\Needspace{14\baselineskip}
\subsection{Case F2: Pseudo-quantification from ungrounded coordinates}\label{app:pseudo_quantification}

\begin{casecard}{F2 summary: spatial and parametric action grounding}
\small
\begin{tabularx}{\linewidth}{@{}>{\bfseries}p{0.20\linewidth}X@{}}
Clinical objective & Longitudinal response assessment between week-040 and week-055 for Patient-059. \\
Ground truth / prediction & RANO partial response (PR) / predicted progressive disease (PD). \\
Trace signal & The agent converted weak screen-space boxes into RAS coordinates and then reported millimeter-scale lesion diameters and SPD change. \\
Why it matters & The reported quantitative evidence was not grounded in validated lesion boundaries, a segmentation mask, a registered longitudinal frame, or a measurement procedure. \\
\end{tabularx}
\end{casecard}

\begin{tracecompact}{Minimal auditable trace}
# The agent selected two axial offsets from different timepoints.
FOLLOWUP CT1 axial offset = 58.159 mm
BASELINE CT1 axial offset     = 44.529 mm

# The same screen-space box was converted into RAS coordinates.
image_coords_to_ras on FOLLOWUP CT1:
  bbox_1000  = [640, 320, 860, 540]
  center_ras = [-25.55, 11.55, 58.159]

image_coords_to_ras on BASELINE CT1:
  bbox_1000  = [640, 320, 860, 540]
  center_ras = [-16.64, -15.54, 44.529]

# These derived coordinates were elevated into quantitative evidence.
BASELINE lesion:
  reported ~= 13.5 mm x 12.3 mm
  SPD      ~= 166 mm^2

FOLLOWUP lesion:
  reported ~= 18.3 mm x 19.9 mm
  SPD      ~= 364 mm^2

final rationale:
  "approximately 119% SPD increase" -> progressive disease
\end{tracecompact}

\begin{mechanismbox}
\small
\textbf{Mechanism.} The measurements were not produced by a validated measurement workflow. They were derived from weakly grounded screen-space boxes, converted into RAS coordinates, and interpreted as lesion diameters without checking the lesion boundary, slice plane, or longitudinal frame.

\textbf{Benchmark implication.} This exposes a spatial-grounding failure: uncertain visual localization was transformed into precise millimeter-scale clinical evidence. The failure is not merely an inaccurate size estimate; it is the invalid elevation of weak evidence into quantitative RANO-style reasoning.
\end{mechanismbox}

\Needspace{14\baselineskip}
\subsection{Case F3: Transform-layer misbinding in registered MRI fusion}\label{app:transform_layer_misbinding}

\begin{casecard}{F3 summary: stateful artifact management}
\small
\begin{tabularx}{\linewidth}{@{}>{\bfseries}p{0.20\linewidth}X@{}}
Clinical objective & Longitudinal response assessment between week-112 and week-152 for Patient-029. \\
Ground truth / prediction & RANO complete response (CR) / predicted stable disease (SD). \\
Trace signal & The agent created separate CT1 and FLAIR transforms, but the executed trace resolved the CT1 transform application to the FLAIR transform and attached both FOLLOWUP volumes to the wrong transform layer. \\
Why it matters & The final viewer state displayed FLAIR fusion, while the final rationale cited registered CT1 evidence. \\
\end{tabularx}
\end{casecard}

\begin{tracecompact}{Minimal auditable trace}
# Two transform objects are created.
brains_fit:
  moving = FOLLOWUP__week-152__CT1
  fixed  = BASELINE__week-112__CT1
  output = FOLLOWUP_CT1_registered_to_BASELINE_CT1

brains_fit:
  moving = FOLLOWUP__week-152__FLAIR
  fixed  = BASELINE__week-112__FLAIR
  output = FOLLOWUP_FLAIR_registered_to_BASELINE_FLAIR

# Raw model call attempts to apply the CT1 transform to CT1.
tool_call_start:
  apply_transform(
    node_ref      = FOLLOWUP__week-152__CT1,
    transform_ref = FOLLOWUP_CT1_registered_to_BASELINE_CT1
  )

# Executed trace resolves the operation to the FLAIR transform.
tool_trace:
  resolved transform = FOLLOWUP_FLAIR_registered_to_BASELINE_FLAIR
  transformed nodes  = [
    FOLLOW_UP__week-152__CT1,
    BASELINE__week-152__FLAIR
  ]

# list_transforms exposes the mismatch.
FOLLOWUP_CT1_registered_to_BASELINE_CT1
  -> []

FOLLOW_UP__FLAIR_registered_to_BASELINE_FLAIR
  -> [
       FOLLOW_UP__week-152__CT1,
       FOLLOW_UP__week-152__FLAIR
     ]

# Viewer state and final rationale disagree.
viewer fusion:
  background = BASELINE__week-112__FLAIR
  foreground = FOLLOW_UP__week-152__FLAIR
  opacity    = 0.5

final rationale:
  "Axial and coronal views of the registered CT1 volumes show..."
\end{tracecompact}

\begin{mechanismbox}
\small
\textbf{Mechanism.} The agent failed to maintain a consistent binding between transform objects, image nodes, viewer layers, and textual evidence. The CT1 transform was left without the expected CT1 child node, while the FLAIR transform contained both the FOLLOWUP CT1 and FOLLOWUP FLAIR nodes.

\textbf{Benchmark implication.} This exposes a software-state failure. The registration algorithm produced plausible objects, but the agent did not verify whether the displayed layers matched the evidence later cited in the final rationale.
\end{mechanismbox}

\Needspace{14\baselineskip}
\subsection{Case F4: Self-confirming segmentation after discarding a correct candidate}\label{app:self_confirming_segmentation}

\begin{casecard}{F4 summary: operation-result calibration and verification}
\small
\begin{tabularx}{\linewidth}{@{}>{\bfseries}p{0.20\linewidth}X@{}}
Clinical objective & Lung PET/CT structured prediction for case R01-004. \\
Ground truth / prediction & \texttt{1D2A3C4B5B} / predicted \texttt{1C2B3A4A5B}. Only the histopathological grade answer was correct. \\
Trace signal & An early VISTA3D label-prompt segmentation~\citep{he2025vista3d} was close to the reference tumor, but the agent discarded it and later reinforced an incorrect manually selected candidate point. \\
Why it matters & The operation sequence converted a mistaken localization into a new segmentation artifact and then into final coordinate evidence. \\
\end{tabularx}
\end{casecard}

\begin{tracecompact}{Minimal auditable trace}
# Reference evidence location.
GT center_ras ~= [-65.78, 58.50, -132.25]
GT representative_point_ras ~= [-65.38, 58.80, -131.75]

# Early VISTA3D output is close to the reference.
vista3d_segment:
  label_prompt = [23]
  segmentation = VISTA3D 23

center_on_segment:
  center_ras = [-65.38, 58.80, -130.50]

# The agent discards the useful candidate and selects a different point.
set_slice_offset:
  axial offset = -265.86

image_coords_to_ras:
  point_1000 = [250, 450]
  point_ras  = [100.60, 46.88, -265.86]

# The incorrect point is used to prompt another segmentation.
vista3d_segment:
  label_prompt = [23]
  points_ras   = [[100.598, 46.881, -265.86]]

center_on_segment:
  center_ras = [102.73, 44.72, -265.50]

# Final answer adopts the later, incorrect localization.
KEY_SLICE_AXIAL_MM:    -265.5
KEY_SLICE_SAGITTAL_MM: 102.7
KEY_SLICE_CORONAL_MM:  44.7
RAS: [102.7, 44.7, -265.5]
\end{tracecompact}

\begin{mechanismbox}
\small
\textbf{Mechanism.} The segmentation operation was not uniformly unreliable. The early label-prompt segmentation was close to the ground truth, but the agent failed to preserve it as primary evidence. A later manual point selection produced an incorrect candidate, and the subsequent point-prompt segmentation reinforced that wrong hypothesis.

\textbf{Benchmark implication.} This exposes an operation-calibration failure: the agent did not arbitrate between competing operation outputs or verify anatomical plausibility before using an operation artifact as final coordinate evidence.
\end{mechanismbox}

\FloatBarrier
\subsection{Instability in Agentic Use of Medical Imaging Software}\label{app:tool_use_instability}

The preceding failure cases analyze individual traces. Here we test a different property: whether the same agent can repeat a professional imaging workflow consistently under identical case settings. The answer is currently no. These runs do not primarily evaluate the underlying registration or segmentation algorithms; they evaluate whether the agent can compose them into a stable, clinically meaningful procedure.

\begin{figure*}[t]
    \centering
    \includegraphics[width=1\linewidth]{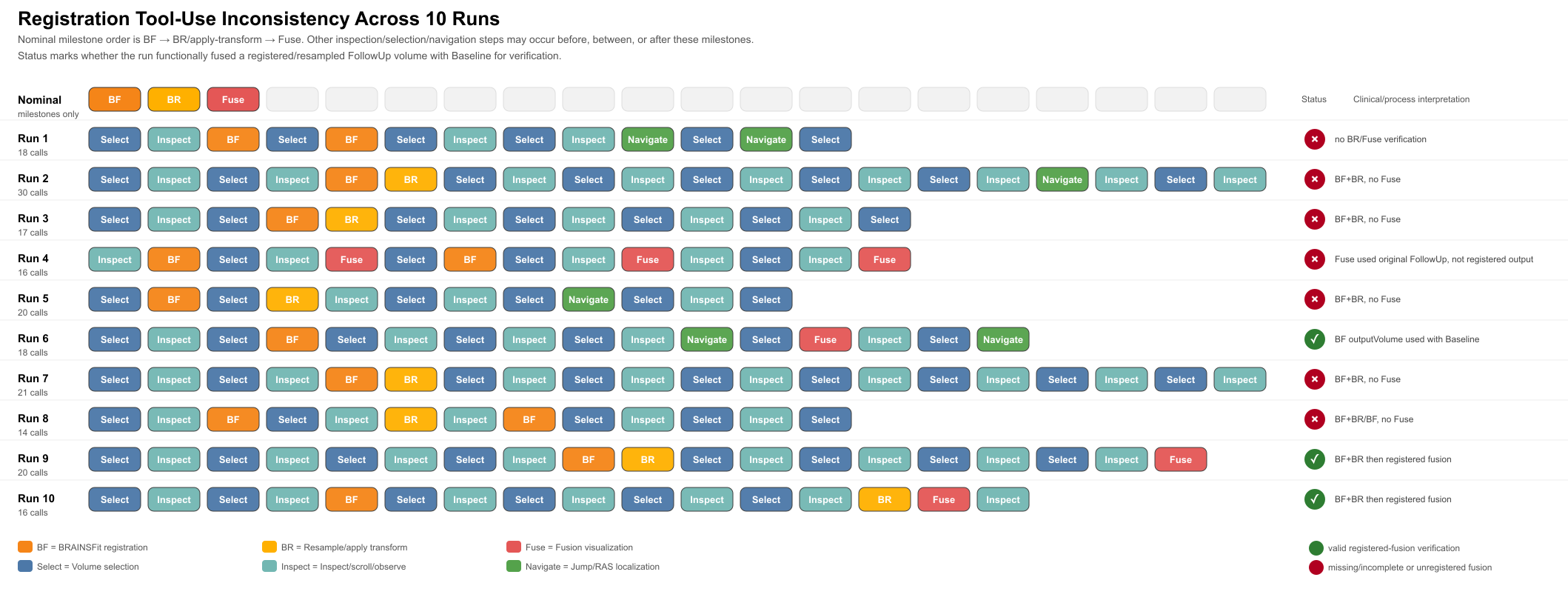}
    \caption{\textbf{Procedural inconsistency in agent-driven registration across 10 repeated runs.} The nominal milestone order is BRAINSFit registration (BF), resample/apply transform (BR), and registered fusion verification. Green status marks runs that complete a functionally valid registered-fusion workflow; red status marks missing, incomplete, or unregistered fusion. Only 3 of 10 runs complete the expected sequence, showing that the agent's operation calls are not reliably organized into a clinically valid registration workflow.}
    \label{fig:registration_workflow_inconsistency}
\end{figure*}

For registration, the nominal workflow is BRAINSFit registration (BF)~\citep{johnson2007brainsfit}, followed by BRAINSResample or apply-transform (BR), followed by registered follow-up versus baseline fusion. Figure~\ref{fig:registration_workflow_inconsistency} shows that only 3 of 10 runs complete this functional sequence. Several trajectories execute BF and BR but never verify with fusion; some call fusion on an unregistered follow-up volume; others repeat selection or inspection calls without closing the loop. This is a procedural-control failure: the agent can invoke the correct named operations but does not reliably maintain the required operation order, artifact binding, and verification step.

\begin{figure*}[t]
    \centering
    \includegraphics[width=1\linewidth]{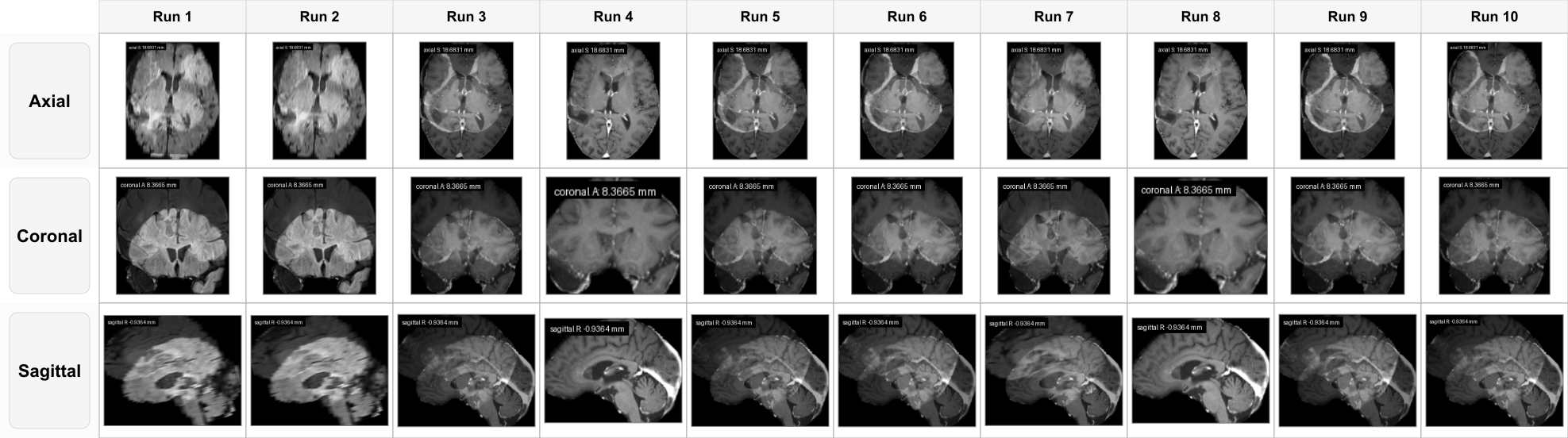}
    \caption{\textbf{Final registration visualizations from 10 repeated runs of the same case.} Each column shows the final axial, coronal, and sagittal views produced by one run. The views differ despite identical inputs and prompts, indicating that procedural instability affects not only the call trace but also the displayed image evidence. This undermines the assumption that access to a registration operation automatically yields a stable, auditable comparison.}
    \label{fig:registration_10_runs_comparison}
\end{figure*}

The visual consequence is shown in Figure~\ref{fig:registration_10_runs_comparison}. Even with identical inputs, the final axial, coronal, and sagittal views differ markedly across runs. Some outputs appear to show plausible registered fusion, while others display inconsistent contrast, missing overlay behavior, or different effective source layers. Thus, workflow instability propagates into the visual evidence that the final answer would cite.

\begin{figure*}[t]
    \centering
    \includegraphics[width=1\linewidth]{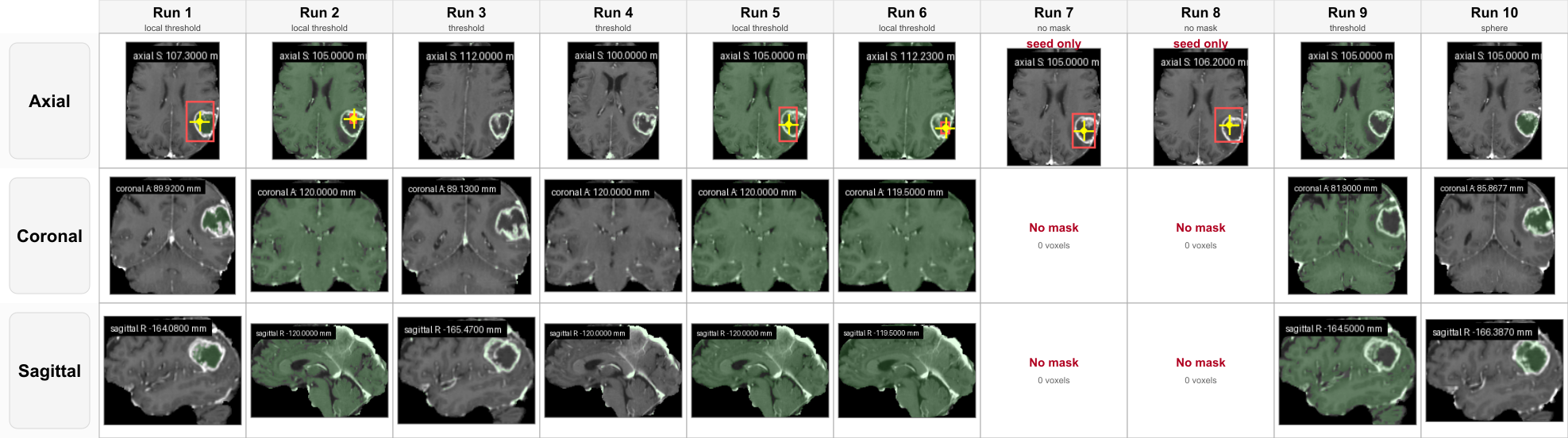}
    \caption{\textbf{Final segmentation outputs from 10 repeated runs of the same case.} Green overlays show the masks generated through the segmentation operation; yellow markers and red boxes show selected seed or candidate regions. Across identical runs, outputs range from broad over-segmentation to local masks, seed-only evidence, and no-mask failures. This illustrates unstable agentic control of segmentation parameters, source state, and failure recovery.}
    \label{fig:segmentation_10_runs_comparison}
\end{figure*}

Segmentation shows the same problem in a different form. Figure~\ref{fig:segmentation_10_runs_comparison} shows that repeated runs produce substantially different masks: some cover broad non-lesion brain tissue, some produce local tumor-like overlays, some return seed-only evidence, and two runs return no mask. The issue is not just mask quality; it is unstable agentic use of the segmentation operation, including different prompt types, seed choices, thresholds, and failure recovery behavior.

\FloatBarrier


\lstdefinestyle{medprompt}{
  basicstyle=\ttfamily\scriptsize,
  breaklines=true,
  columns=fullflexible,
  keepspaces=true,
  showstringspaces=false,
  frame=single,
  framerule=0.25pt,
  linewidth=\dimexpr\linewidth-1.0em\relax,
  xleftmargin=0.8em,
  xrightmargin=0.2em,
  framexleftmargin=0.4em,
  framexrightmargin=0.4em,
  framextopmargin=0.25em,
  framexbottommargin=0.25em,
  aboveskip=0.6\baselineskip,
  belowskip=0.6\baselineskip
}

\providecommand{\promptblockheading}[1]{%
  \Needspace{7\baselineskip}%
  \paragraph{#1}\leavevmode\par\nobreak\smallskip\noindent
}

\section{Prompt and Tool Interface Details}
\label{app:prompt_tool_interface}

This appendix documents the prompt templates and model-visible tools used by \bench{}.  Placeholders such as \texttt{<CASE\_ID>}, \texttt{<SCENE\_INVENTORY>}, and \texttt{<CURRENT\_VIEWER\_SCREENSHOT>} indicate text or images filled separately for each case before the model sees the prompt.  The reported benchmark scores do not use human or LLM judging; answers are scored by deterministic parsing and hidden-reference evidence checks.  Full case-level prompts, screenshots, tool-call traces, tool outputs, viewer states, and final responses are released with the code artifact.

\paragraph{Disclosure granularity.}
The appendix includes the system-level task instructions, answer formats, evidence formats, coordinate conventions, tool budget, and callable tool names exposed to the model.  It does not enumerate every case-instantiated prompt in the PDF because those prompts repeat the same templates and differ mainly in case identifiers, loaded image names, scene inventories, and attached screenshots.

\begin{table*}[t]
\centering
{\scriptsize
\setlength{\tabcolsep}{3.0pt}
\renewcommand{\arraystretch}{1.12}
\begin{adjustbox}{max width=\textwidth}
\begin{tabularx}{\textwidth}{@{}p{0.98in}p{0.82in}X X p{0.72in}@{}}
\toprule
Task family & Software & Template included below & Final output & Budget \\
\midrule
Longitudinal MRI (LUMIERE) & 3D Slicer & Radiology system instruction, OLD/CURRENT study template, RANO options, CURRENT-study evidence fields & RANO option plus six CURRENT-study YES/NO evidence fields & 20 rounds \\
Brain MRI (UCSF-PDGM) & 3D Slicer & Multi-sequence MRI instruction, diagnostic options, tumor-evidence template when evidence localization is evaluated & Diagnostic option; key slices and RAS point when evidence localization is evaluated & 20 rounds \\
Lung CT/PET (NSCLC) & 3D Slicer & Thoracic CT/PET instruction, series-catalog placeholder, five-question MCQ schema, lesion-evidence template when evidence localization is evaluated & Compact five-question answer string; key slices, RAS point, and source-series provenance when evidence localization is evaluated & 20 rounds \\
Breast WSI (BRACS) & QuPath & QuPath system instruction, slide-level task prompt, per-round observation template, JSON output schema & JSON with BRACS label, confidence, rationale, and ROI coordinates for non-normal predictions & 20 rounds \\
Lymph-node WSI (CAMELYON17) & QuPath & QuPath system instruction, slide-level tumor task, per-round observation template, JSON output schema & JSON with tumor presence, largest metastasis category, confidence, rationale, and tumor coordinates for positive predictions & 20 rounds \\
\bottomrule
\end{tabularx}
\end{adjustbox}}
\caption{Prompt templates and answer formats disclosed for each benchmark module.  Case-specific text and images are represented by placeholders and filled before inference.}
\label{tab:prompt_components}
\end{table*}

\subsection{3D Slicer Radiology Prompt Templates}
\label{app:slicer_prompt_templates}

For radiology episodes, the system preloads the relevant study or studies and attaches a scene inventory plus baseline axial, sagittal, and coronal screenshots after the text prompt.  LUMIERE, UCSF-PDGM, and NSCLC use the same viewer-native 3D Slicer tool surface and the same 20-round budget.  Basic foreground/background fusion of already loaded or prealigned volumes is part of this shared surface.

\paragraph{Longitudinal MRI response assessment.}
The template below is used for LUMIERE longitudinal response-assessment episodes. The agent compares matched modalities across timepoints and reports evidence fields for the CURRENT study only.

\begin{lstlisting}[style=medprompt]
System/task instruction:
You are an AI radiology assistant using a remote MRI viewer (3D Slicer). The scene already contains two MRI studies from the same patient: OLD and CURRENT. Do not call open_case unless the viewer is clearly empty or broken. If recovery is needed, use the exact paths listed in the user message so both OLD and CURRENT studies are loaded together. The scene inventory is provided in the prompt. Use the viewer-native 3D Slicer tools exposed in this run. Focus on the CURRENT study; OLD is only the reference for interval comparison. Compare matching modalities across OLD and CURRENT. CT1 is usually the most important sequence for RANO response assessment, with FLAIR/T2/T1 as supporting context. Choose exactly one of the four RANO response classes. Also output the requested YES/NO evidence fields for the CURRENT study only. Do not output UNKNOWN and do not output JSON. End with the structured block below, with ANSWER as the final line.

Tool-use rule:
When calling image_coords_to_ras, send exactly one of bbox_1000 or point_1000, never both. Use bbox_1000 for a box center and point_1000 for a clicked point.

Tool budget:
At most 20 tool-calling rounds. If the budget is exhausted, immediately provide the best final answer using the evidence already in context.

Case brief:
Case brief for <PATIENT_ID>__week-OLD__to__week-CURRENT:
- Dataset: LUMIERE
- Image space: DeepBraTumIA atlas-space skull-stripped MRI (MNI)
- Patient: <PATIENT_ID>
- OLD reference study: week-OLD
- CURRENT study to classify: week-CURRENT
- Loaded modalities for both studies: CT1, T1, T2, FLAIR
- Tumor segmentation masks are intentionally hidden.
- Evidence fields describe the CURRENT study only.

Task:
Compare the CURRENT study against the OLD reference study and choose the CURRENT study's RANO response label.
Answer options:
  A) PD
  B) SD
  C) PR
  D) CR

After a short rationale, end with exactly this block:
NON_MEASURABLE_LESION: <YES|NO>
T2_PROGRESSION: <YES|NO>
CURRENT_TARGET_LESION: <YES|NO>
NEW_TARGET_LESION: <YES|NO>
CURRENT_NON_TARGET_LESION: <YES|NO>
NEW_NON_TARGET_LESION: <YES|NO>
ANSWER: <A|B|C|D>
\end{lstlisting}

\paragraph{Multi-sequence brain MRI diagnosis.}
The UCSF-PDGM template uses the same viewer-native 3D Slicer tool surface.  When tumor localization is evaluated, the model localizes the lesion in shared RAS space using an informative loaded anatomical MRI sequence.

\begin{lstlisting}[style=medprompt]
System/task instruction:
You are an AI radiology assistant using a remote MRI viewer (3D Slicer). The case has already been opened by the system before you start. Do not call open_case unless the viewer is clearly empty or broken. If recovery is needed, use the exact absolute path provided in the user message. The case is opened with a core multi-sequence MRI subset, and the current scene inventory is provided in the prompt. Use the viewer-native 3D Slicer tools exposed in this run. If KEY_SLICE_*_MM evidence is requested, report canonical RAS-axis millimeters from the image corner overlay: axial=S, sagittal=R, coronal=A. Do not use slice indices. When calling image_coords_to_ras, send exactly one of bbox_1000 or point_1000. Do not output JSON. If evidence is requested, include it exactly once before the final answer line. Output must contain a final line exactly like: ANSWER: <A|B|C|D|E|F>.

Tool budget:
At most 20 tool-calling rounds. If the budget is exhausted, immediately provide the best final answer using the evidence already in context.

Case brief:
Case brief for <CASE_ID>:
- CSV ID: <CSV_ID>
- Loaded core NIfTI files: <comma-separated core T1/T1c/T2/FLAIR-style files>
- The current scene inventory is shown after open_case.
- Treat the loaded T1/T1c/T2/FLAIR-style sequences as the main evidence sources.
- If tumor localization is evaluated, evidence may be measured on any loaded anatomical MRI sequence where the tumor is clearly visible.

Task:
Classify the case into one option:
  A) Glioblastoma
  B) Oligodendroglioma / Astrocytoma
  C) No tumor
  D) Dysembryoplastic neuroepithelial tumor (DNET) / Ganglioglioma
  E) Pilocytic Astrocytoma
  F) Central Neurocytoma / Ependymoma

Suggested steps:
1) Review baseline views.
2) Use the prompt-provided scene inventory to choose among loaded sequences; use select_volume to switch sequences.
3) Use observe/scroll/zoom/wl as needed. Prefer scroll output='keyframes' unless video is necessary.

If evidence is requested, output exactly once before the answer line:
EVIDENCE:
KEY_SLICE_AXIAL_MM: <float>
KEY_SLICE_SAGITTAL_MM: <float>
KEY_SLICE_CORONAL_MM: <float>
RAS: [<R>, <A>, <S>]
Definitions:
- KEY_SLICE_*_MM values use canonical RAS-axis millimeters from the image corner overlay: axial=S, sagittal=R, coronal=A.
- RAS must be a single point strictly inside the visible tumor mask.
- Use one loaded MRI sequence consistently while localizing the tumor and producing the evidence block.
- If image_coords_to_ras is used, pass exactly one of bbox_1000 or point_1000.

Final line:
ANSWER: <A|B|C|D|E|F>
\end{lstlisting}

\paragraph{Lung CT/PET structured prediction.}
The NSCLC prompt asks for five structured predictions.  When lesion localization is evaluated, the evidence block must be produced on the specified loaded source series.

\begin{lstlisting}[style=medprompt]
System/task instruction:
You are a clinically grounded multimodal thoracic imaging assistant using a remote 3D Slicer viewer. All resolved imaging series across available timepoints are already loaded, and the current scene inventory is provided. Use CT morphology first. Use PET as adjunctive evidence for metabolic activity and nodal survey. For chest CT inspection, use window_preset with ct_lung and ct_mediastinal as needed. If KEY_SLICE_*_MM evidence is requested, report canonical RAS-axis millimeters from the image corner overlay: axial=S, sagittal=R, coronal=A. Do not use the native viewer-header sign convention if it differs from the corner overlay. Use the viewer-native 3D Slicer tools exposed in this run. When calling image_coords_to_ras, send exactly one of bbox_1000 or point_1000. Do not output JSON. Write a brief English rationale. If evidence is requested, include the evidence block exactly as specified before the final compact answer string. End with a single compact answer string exactly like 1C2D3B4A5B, and do not repeat it.

Tool budget:
At most 20 tool-calling rounds. If the budget is exhausted, immediately provide the best final answer using the evidence already in context.

Case brief:
Case brief for <CASE_ID>:
- All resolved CT/PET series across available timepoints were preloaded at start.
- Preferred initial viewing heuristic: <DEFAULT_ALIAS, usually diagnostic_ct>
- Dynamic CT/PET series catalog, default aliases, loaded series names, and curated grouping information are inserted here.
- If lesion localization is evaluated, the evidence source series is <SOURCE_VOLUME_NAME> (<SOURCE_MODALITY>). The model must switch/select this loaded series before measuring KEY_SLICE_*_MM or RAS and must keep it active while calling image_coords_to_ras.

Task:
Answer the following five multiple-choice questions based on the imaging only. Return the final answer in one compact string, with no spaces, exactly like 1C2D3B4A5B.

Q1. Tumor location
1A Right upper lobe; 1B Right middle lobe; 1C Right lower lobe; 1D Left upper lobe; 1E Left lower lobe; 1F Lingula; 1G No tumor

Q2. Pathological T stage
2A T1; 2B T2; 2C T3; 2D T4; 2E Not available / not collected; 2F No tumor

Q3. Pathological N stage
3A N0; 3B N1; 3C N2; 3D Not available / not collected; 3E No tumor

Q4. Histology
4A Adenocarcinoma; 4B Squamous cell carcinoma; 4C Small cell carcinoma; 4D Other / indeterminate NSCLC; 4E No tumor

Q5. Histopathological grade
5A G1 Well differentiated; 5B G2 Moderately differentiated; 5C G3 Poorly differentiated; 5D Other Type I: well to moderately differentiated; 5E Other Type II: moderately to poorly differentiated; 5F No tumor

If evidence is requested, output exactly once before the compact answer string:
EVIDENCE:
KEY_SLICE_AXIAL_MM: <float>
KEY_SLICE_SAGITTAL_MM: <float>
KEY_SLICE_CORONAL_MM: <float>
RAS: [<R>, <A>, <S>]
Definitions:
- The evidence must be reported on the loaded source series <SOURCE_VOLUME_NAME> (<SOURCE_MODALITY>) only.
- Each KEY_SLICE_*_MM value uses canonical RAS-axis millimeters from the image corner overlay on the active source series: axial=S, sagittal=R, coronal=A.
- RAS must be a single point strictly inside the visible primary tumor on that same source series.
- If image_coords_to_ras is used, pass exactly one of bbox_1000 or point_1000.
- If needed, explicitly call select_volume(volume_name=...) on the source series before the evidence block.
\end{lstlisting}

\subsection{QuPath Pathology Prompt Templates}
\label{app:qupath_prompt_templates}

The pathology setting operates in QuPath through a direct tool interface.  The model receives a task prompt, a direct QuPath state summary, a rule-based GUI/control summary, and the current viewer screenshot at each step.  The screenshot is the primary image evidence; the state summaries are software-state evidence, not pathology evidence.  Pathology episodes use a 20-round budget, with at most one tool call per round.

\promptblockheading{Common QuPath system instruction.}
\begin{lstlisting}[style=medprompt]
You are the host model being evaluated. You operate QuPath through an explicit, audited tool surface.

Core rules:
- You are the primary agent. Do not delegate the core reasoning task.
- The attached viewer screenshot is the primary evidence for pathology interpretation.
- The direct QuPath state summary reports structured software/viewer state from QuPath itself.
- The GUI state summary is rule-based and is not pathology evidence.
- Prefer direct tools whenever they can perform the needed viewer action.
- Work inside QuPath only. Do not assume access to the full computer.
- Use at most one tool call per step.
- Use at most 20 tool-calling rounds.
- If the task is complete, answer directly without calling a tool.
- The case is usually already loaded before the first step. Do not call load_case unless recovery or reset is actually needed.
- Never invent GUI state, direct QuPath state, or pathology findings that are not grounded in the current screenshot or tool outputs.
- Never claim a requested viewer magnification was reached unless the current direct QuPath state summary confirms it.
\end{lstlisting}

\promptblockheading{BRACS slide-level prompt.}
\begin{lstlisting}[style=medprompt]
Case summary:
Case alias: <CASE_ALIAS>
A blinded breast H&E whole-slide image is already loaded in QuPath.
Ground-truth annotations are hidden from you.
You must determine the single best slide-level BRACS label and, when non-negative, support it with one or more WSI coordinate localizations.

Task:
Choose the single best slide-level BRACS label:
A. N
B. PB
C. UDH
D. FEA
E. ADH
F. DCIS
G. IC

Rules:
- The image is already loaded in QuPath and ground-truth annotations are hidden.
- If the final slide-level label is N, localizations must be an empty list.
- If the final label is not N, output one or more ROI coordinates in full-resolution WSI image pixels.
- To obtain a coordinate, call viewer_point_to_wsi_coordinate with x and y in 0..999 on the current viewer screenshot, measured from the top-left corner.
- Use the x and y returned by that tool in the final answer.
- Choose points on the main tissue image, not on the overview thumbnail, scale bar, or text overlays.
- wsi_label must be exactly one of: N, PB, UDH, FEA, ADH, DCIS, IC.
- If uncertain, state uncertainty plainly; do not invent findings.

Return JSON only, with exactly these keys:
{
  "selected_option": "A|B|C|D|E|F|G",
  "wsi_label": "N|PB|UDH|FEA|ADH|DCIS|IC",
  "confidence": 0.0,
  "brief_rationale": "...",
  "localizations": [{"x": 0.0, "y": 0.0}]
}
\end{lstlisting}

\promptblockheading{CAMELYON17 slide-level prompt.}
\begin{lstlisting}[style=medprompt]
Case summary:
Case alias: <CASE_ALIAS>
A blinded CAMELYON17 lymph node H&E whole-slide image is already loaded in QuPath.
Ground-truth masks and annotations are hidden from you.
This run is slide-level only. Determine whether tumor is present, select the slide-level metastasis category, and output tumor coordinates when positive.

Task:
1) Tumor presence?
A. no_tumor
B. tumor_present

2) Largest metastasis category for this slide?
A. negative
B. itc
C. micro
D. macro

Rules:
- The image is already loaded in QuPath; ground-truth masks and annotations are hidden.
- This run is slide-level only. Do not answer patient-level staging questions.
- If tumor is absent and the slide-level label is negative, localizations must be an empty list.
- If tumor is present, output one or more tumor coordinates in full-resolution WSI image pixels.
- To obtain a coordinate, call viewer_point_to_wsi_coordinate with x and y in 0..999 on the current viewer screenshot, measured from the top-left corner.
- Use the x and y returned by that tool in the final answer.
- Choose points on the main tissue image, not on the overview thumbnail, scale bar, or text overlays.
- Prefer relative zooming and visible tissue context over trusting nominal magnification metadata alone.
- If a zoomed-in view lands on blank background or off-tissue area, zoom back out or pan until informative tissue is visible again.
- If uncertain, state uncertainty plainly; do not invent findings.

Return JSON only, with exactly these keys:
{
  "tumor_presence_option": "A|B",
  "tumor_presence": "no_tumor|tumor_present",
  "largest_metastasis_option": "A|B|C|D",
  "largest_metastasis": "negative|itc|micro|macro",
  "confidence": 0.0,
  "brief_rationale": "...",
  "localizations": [{"x": 0.0, "y": 0.0}]
}
\end{lstlisting}

\promptblockheading{Per-round QuPath observation template.}
\begin{lstlisting}[style=medprompt]
Task:
<BENCHMARK_TASK_PROMPT>

Case summary:
<CASE_SUMMARY_TEXT>

Current QuPath direct state summary (software/viewer state from QuPath itself):
<CURRENT_QUPATH_DIRECT_STATE_SUMMARY>

Current QuPath GUI state summary (rule-based GUI/control state, not pathology evidence):
<CURRENT_QUPATH_RULE_BASED_GUI_STATE_SUMMARY>

Current viewer evidence:
- The current QuPath viewer screenshot is attached in this message.
- Use the attached screenshot itself for pathology or image-content reasoning.

Decide the next best action:
- either call exactly one tool
- or provide the final answer
\end{lstlisting}

\subsection{Model-Visible Tools}
\label{app:tool_surfaces}

The runtime exposes named functions only.  Raw Python execution in 3D Slicer and arbitrary script execution in QuPath are not provided to the evaluated agent.  Every call is logged with its arguments, returned state, screenshots or artifacts, and final answer text.

\begin{table*}[t]
\centering
{\scriptsize
\setlength{\tabcolsep}{2.5pt}
\renewcommand{\arraystretch}{1.08}
\begin{adjustbox}{max width=\textwidth}
\begin{tabularx}{\textwidth}{@{}p{1.35in}p{1.95in}X@{}}
\toprule
Tool & Arguments & Purpose and returned information \\
\midrule
\texttt{ping} & -- & Checks whether the Slicer WebServer is responding. \\
\texttt{open\_case} & \texttt{dicom\_dir}; \texttt{spacing}; \texttt{slice\_planes}; \texttt{view\_layout}; \texttt{size} & Opens or recovers a case path with the standard viewer preset; returns baseline views and scene inventory. \\
\texttt{select\_volume} & \texttt{volume\_id}; \texttt{volume\_name}; \texttt{view}; \texttt{size} & Selects an already loaded image volume and returns recovered standard views; used for switching MRI sequences or CT/PET series. \\
\texttt{observe} & \texttt{view}; \texttt{size}; \texttt{include\_metadata} & Returns the current axial, sagittal, and coronal views, or a selected view. \\
\texttt{scroll} & \texttt{view}; \texttt{direction}; \texttt{num\_slices}; \texttt{step}; \texttt{output}; \texttt{size} & Scrolls through slices and returns sampled keyframes or a cine. \\
\texttt{zoom} & \texttt{view}; \texttt{factor}; \texttt{center\_1000}; \texttt{size} & Zooms a slice view relative to its current field of view and returns the updated view. \\
\texttt{fit} & \texttt{view}; \texttt{size} & Fits the selected slice view to the full volume and returns the updated view. \\
\texttt{set\_slice\_offset} & \texttt{view}; \texttt{offset}; \texttt{size} & Sets an axial, sagittal, or coronal slice position directly in millimeters and returns the updated view. \\
\texttt{get\_slice\_offset\_range} & \texttt{view} & Returns the valid minimum and maximum offset range for the selected view. \\
\texttt{recover\_standard\_views} & \texttt{volume\_id}; \texttt{volume\_name}; \texttt{size} & Restores axial, sagittal, and coronal views around a volume center. \\
\texttt{image\_coords\_to\_ras} & \texttt{view}; \texttt{point\_1000}; \texttt{bbox\_1000}; \texttt{volume\_id}; \texttt{volume\_name} & Converts normalized screen coordinates to physical RAS millimeters for the active loaded volume. \\
\texttt{get\_viewer\_state} & \texttt{include\_volumes}; \texttt{include\_fusion}; \texttt{include\_slice\_positions} & Returns active volume, slice offsets, fusion state, and related viewer state. \\
\texttt{wl} & \texttt{view}; \texttt{window}; \texttt{level}; \texttt{auto}; \texttt{volume\_id}; \texttt{volume\_name}; \texttt{size} & Sets window/level and returns the updated view. \\
\texttt{window\_preset} & \texttt{preset}; \texttt{view}; \texttt{volume\_id}; \texttt{volume\_name}; \texttt{size} & Applies modality presets such as \texttt{ct\_lung}, \texttt{ct\_mediastinal}, \texttt{ct\_bone}, or \texttt{pet\_auto}; returns the updated view. \\
\texttt{jump} & \texttt{ras}; \texttt{view}; \texttt{size} & Jumps linked views to a known RAS point and returns the updated views. \\
\texttt{set\_layout} & \texttt{layout}; \texttt{size} & Sets a standard Slicer layout and returns updated views. \\
\texttt{fusion} & \texttt{background\_volume\_ref}; \texttt{foreground\_volume\_ref}; \texttt{foreground\_opacity}; \texttt{mode}; \texttt{view}; \texttt{size} & Sets basic foreground/background slice fusion for already loaded or prealigned volumes, e.g., CT+PET, and returns the updated fused view. \\
\bottomrule
\end{tabularx}
\end{adjustbox}}
\caption{Model-visible viewer-native 3D Slicer tools used by all radiology benchmark modules.  The released schemas contain the exact JSON definitions used at runtime.}
\label{tab:slicer_core_tools}
\end{table*}

\begin{table*}[t]
\centering
{\scriptsize
\setlength{\tabcolsep}{3.0pt}
\renewcommand{\arraystretch}{1.08}
\begin{adjustbox}{max width=\textwidth}
\begin{tabularx}{\textwidth}{@{}p{1.35in}p{1.75in}X@{}}
\toprule
Tool & Arguments & Purpose and returned information \\
\midrule
\texttt{zoom} & \texttt{direction}; \texttt{amount}; \texttt{unit}; \texttt{x}; \texttt{y} & Changes the QuPath viewer zoom by absolute magnification, relative factor, or internal steps; can recenter on a normalized screenshot point. \\
\texttt{pan} & \texttt{direction}; \texttt{amount}; \texttt{unit}; \texttt{x}; \texttt{y} & Pans the QuPath viewer or recenters the view; returns the updated viewer state. \\
\texttt{focus\_qupath} & -- & Brings QuPath to the foreground and returns the current viewer state. \\
\texttt{capture\_viewer} & \texttt{label} & Captures the current viewer and stores a labeled screenshot artifact. \\
\shortstack[l]{\texttt{viewer\_point\_to}\\\texttt{\_wsi\_coordinate}} & \texttt{x}; \texttt{y} & Converts a normalized point on the current QuPath screenshot to full-resolution WSI pixel coordinates. \\
\bottomrule
\end{tabularx}
\end{adjustbox}}
\caption{Model-visible QuPath direct tools used for BRACS and CAMELYON17.  The released schemas contain the exact JSON definitions used at runtime.}
\label{tab:qupath_tools}
\end{table*}

\paragraph{Coordinate conventions.}
For \texttt{image\_coords\_to\_ras}, screen coordinates are normalized to the range 0--999 from the top-left corner of the displayed slice.  The call uses either a point or a bounding box center, and the returned value is an RAS coordinate in millimeters for the active loaded volume.  For \texttt{viewer\_point\_to\_wsi\_coordinate}, \texttt{x} and \texttt{y} are normalized to 0--999 on the current QuPath screenshot from the top-left corner, and the returned coordinate is in full-resolution WSI pixels.

\subsection{Dynamic Attachments and Logging}
\label{app:dynamic_attachments_logging}

Radiology prompts are followed by a scene-inventory text block and baseline axial, sagittal, and coronal image attachments from 3D Slicer.  Pathology prompts are followed at each step by a QuPath direct state summary, a rule-based GUI/control summary, and the current viewer screenshot.  These attachments are filled separately for each case and therefore are not enumerated in the manuscript.  The runtime stores the instantiated prompt, image attachments, tool-call arguments, tool outputs, viewer states, screenshots, generated artifacts, and final answer for each episode, making the execution reconstructable without relying on model-hidden reasoning.

\FloatBarrier
\section{Licenses and Terms of Use}
\bench{} uses existing open-source software, public research datasets, and public baseline models only under their published licenses and terms of use. We cite the original asset papers and source pages, retain required copyright and license notices, and follow dataset-specific restrictions. In particular, TCIA-hosted datasets are used under the applicable Creative Commons licenses together with the TCIA Data Usage Policy and Restrictions, including the prohibition on attempting to identify or contact data subjects. BRACS and the MONAI/vista3d model weights are used only for non-commercial research/evaluation purposes. Table~\ref{tab:asset_licenses} summarizes the main third-party assets used in the benchmark and the corresponding licenses or usage terms.

\begin{table}[th]
\centering
\small
\caption{Existing assets, licenses, and terms of use.}
\label{tab:asset_licenses}
\begin{tabular}{p{0.23\linewidth} p{0.17\linewidth} p{0.50\linewidth}}
\toprule
Asset & Type & License / terms of use \\
\midrule
3D Slicer & Software & BSD-style license; no restriction on software use. Not FDA approved; no clinical-use claim is made. \\
QuPath & Software & GNU GPL v3.0. Not intended for clinical, diagnostic, or therapeutic use. \\
MONAI & Software & Apache License 2.0. \\
VISTA3D / MONAI VISTA & Model / operation & MONAI VISTA source code is Apache-2.0. The MONAI/vista3d model weights are distributed under the NVIDIA model-weight license for non-commercial research/evaluation use. \\
LUMIERE & Dataset & CC0 public-domain dedication on Figshare; used as de-identified public research data. \\
UCSF-PDGM & Dataset & CC BY 4.0 via TCIA; subject to TCIA Data Usage Policy and Restrictions. \\
NSCLC-Radiogenomics & Dataset & CC BY 3.0 via TCIA; subject to TCIA Data Usage Policy and Restrictions. \\
BRACS & Dataset & Creative Commons Attribution-NonCommercial 4.0 International; used for non-commercial research evaluation with citation. \\
CAMELYON17 & Dataset & CC0 public-domain dedication / open-access challenge data. \\
M3D / RadFM & Baseline models & M3D code is MIT and the evaluated M3D-LaMed checkpoint is listed as Apache-2.0; RadFM code is MIT. Public checkpoints are used only as evaluation baselines under their published project/Hugging Face terms. \\
\bottomrule
\end{tabular}
\end{table}

\end{document}